\documentclass[10pt,twocolumn,letterpaper]{article}

\usepackage{wacv}              % To produce the CAMERA-READY version
\usepackage{amsmath}
\usepackage{amssymb}
\usepackage{graphicx}
\usepackage{booktabs}
\usepackage{multirow}
\usepackage{array}
\usepackage[table,dvipsnames]{xcolor}
\usepackage{longtable}
\usepackage{rotating}    

\usepackage{booktabs,longtable,array,amsmath}
\usepackage[table,dvipsnames]{xcolor}

\definecolor{headerblue}{RGB}{222,235,247}
\definecolor{oursgray}{RGB}{238,238,238}

\usepackage{pifont}
\usepackage{colortbl}
\usepackage{caption}
\usepackage[dvipsnames]{xcolor}
\newcommand{\venue}[1]{\textcolor{NavyBlue}{\scriptsize[#1]}}

\newcommand{\khat}{\hat{k}}

\definecolor{wacvblue}{rgb}{0.21,0.49,0.74}
\usepackage[pagebackref,breaklinks,colorlinks,allcolors=wacvblue]{hyperref}

\def\wacvPaperID{2095} % *** Enter the WACV Paper ID here
\def\confName{WACV}
\def\confYear{2027}

\title{When Can Test-Time Adaptation Help Zero-Shot CT
Vision-Language Models?}

\author{
Ailar Mahdizadeh$^{1,2}$ \quad
Puria Azadi Moghadam$^{1}$ \quad
Xiangteng He$^{1,2}$ \quad
Leonid Sigal$^{1,2}$\\
$^{1}$University of British Columbia \\
$^{2}$Vector Institute \\
\texttt{ailar.mahdizadeh@ubc.ca}
}

\begin{document}
\maketitle

\begin{abstract}
3D CT vision-language models (VLMs) classify abnormalities from text prompts in
a zero-shot manner, which is attractive for cross-institution deployment where
labels are scarce and clinical tasks shift faster than supervised models can be
retrained. A real CT scan, however, typically contains several co-occurring
abnormalities, and the reliability of zero-shot multi-label prediction under
distribution shift remains poorly understood. Test-time adaptation (TTA) can
update a model on unlabeled target scans without source data or target
annotations, yet existing TTA methods target multi-class softmax prediction on
natural images or 2D medical segmentation, and none addresses unsupervised
multi-label adaptation for zero-shot 3D CT VLMs. We study \emph{when} TTA can
help zero-shot 3D CT VLMs, rather than assuming that adaptation is always
beneficial. A controlled diagnostic analysis shows that TTA is conditional. The
volumetric input must preserve the encoder's depth structure, and the base
representation must transfer to the target cohort, with depth reduction alone
lowering internal AUROC by more than $0.12$ before adaptation begins. We then
focus on the regime where the base model already separates present from absent
abnormalities. There, prompt-pair CT diagnosis scores each abnormality as an
independent Bernoulli variable rather than as part of a shared softmax, which
makes standard entropy minimization mismatched to multi-label CT. We introduce
\textbf{CARVE}\footnote{The code is publicly available at: \url{https://github.com/ailarmhz/CARVE.git}.}
(Cardinality-Aware Retained-View Entropy), the first TTA method for this
prompt-pair multi-label regime. CARVE estimates a sample-specific positive-label
cardinality $\khat$, optimizes a top-$\khat$ objective to preserve co-occurring
abnormalities, and performs memory-efficient multi-view adaptation by scoring
weak 3D views without gradients before updating on a retained subset. Across
contrastive CT-CLIP and anatomy-aware fVLM, and across multi-label, three-class,
and binary CT settings, CARVE provides the most consistent improvements when the
base model is already discriminative. These results establish multi-label TTA for
zero-shot 3D CT VLMs as a distinct problem and CARVE as a cardinality-aware
method for it.
\end{abstract}

\section{Introduction}
\label{sec:intro}

3D CT vision-language models, such as CT-CLIP~\cite{hamamci2024ctclip},
fVLM~\cite{shui2025fvlm}, and Merlin~\cite{blankemeier2026merlin}, align
volumetric scans with radiology text and classify abnormalities directly from
text prompts without task-specific labels~\cite{radford2021clip}. This is attractive for deployment
across institutions, where annotation is expensive, label definitions vary, and
new clinical tasks arise faster than supervised models can be trained. The
clinically realistic form of this task is multi-label, since a single CT scan
routinely contains several co-occurring abnormalities, so a deployable model must
score many abnormalities at once rather than choose one. Deployment, however,
also exposes these models to distribution shift, since scans differ across sites
in scanner vendor, reconstruction protocol, demographics, and disease
prevalence~\cite{roschewitz2023automatic,yu2022external}, which can substantially
degrade zero-shot performance.

Distribution shift can be addressed in several ways, which differ mainly in what
they require at deployment. Domain adaptation assumes access to source data
alongside target data~\cite{ganin2016dann}; supervised fine-tuning assumes target labels; domain
generalization intervenes at training time and cannot react to a shift seen only
after deployment. None of these hold for a zero-shot CT VLM moved to a new
institution, where the source corpus is typically unavailable~\cite{liang2020shot}, target abnormality
labels are precisely what is missing, and the shift appears only at inference, one
scan at a time. TTA is appealing because it removes exactly these assumptions, as
it is source-free, label-free, and per-scan~\cite{sun2020ttt,wang2021tent}. The methods that exist, however,
were built for a different setting. TTA for vision-language models targets
multi-class softmax prediction on natural images~\cite{shu2022tpt,wang2021tent,zhao2024test},
multi-label TTA has been studied only on natural-image benchmarks and infers how
many labels are present by retrieving a paired caption~\cite{wu2025mltta}, and
medical TTA has focused mainly on segmentation
benchmarks~\cite{yu2025medsegtta,chen2025grata} or single-label classification
under label-distribution shift~\cite{ma2022ttadc}. None addresses unsupervised
multi-label adaptation for zero-shot 3D CT VLMs. Our question is therefore not
whether to adapt, but \emph{when} adaptation helps in this setting, and with what
objective.

We approach this question diagnostically. Under a single controlled protocol, we
isolate three conditions that determine whether adaptation helps
(\Cref{sec:diagnosis}). The volumetric input must preserve the depth structure
the encoder was pretrained on, the base representation must transfer to the target
cohort, and the adaptation objective must match the prompt-pair multi-label
structure of the prediction. We treat the first two as deployment preconditions
and design a method for the third. That third condition is specific. Prompt-pair
diagnosis scores each abnormality as an independent Bernoulli probability rather
than a shared softmax, so cross-label entropy collapses co-occurring positives
while per-label binary entropy weakens near $p{=}0.5$, leaving one missing
quantity, the number of positives a scan should preserve, which a new objective
can target.

We introduce \textbf{CARVE}, the first TTA method designed for zero-shot
prompt-pair multi-label 3D CT diagnosis. CARVE operates directly on prompt-pair
probabilities, estimates a sample-specific positive label cardinality $\khat$, and
replaces standard entropy minimization with a cardinality-aware top-$\khat$
entropy objective that preserves multiple co-occurring abnormalities. Unlike
multi-label TTA for natural images, which infers the label count by retrieving a
paired caption~\cite{wu2025mltta}, CARVE estimates cardinality directly from the
prompt-pair probabilities, which is what makes it applicable when no caption is
available and volumetric memory rules out dense multi-view backpropagation. To
make adaptation practical for volumetric inference, CARVE scores weak 3D views
without gradients, retains the most reliable subset, and recomputes only those
views for the update. The method adapts a restricted set of visual normalization
parameters and resets episodically for each scan, avoiding drift across the target
stream (\Cref{sec:method}). We evaluate five objectives under a controlled
native-resolution protocol, including No-TTA, TENT~\cite{wang2021tent},
RLCF~\cite{zhao2024test}, ML-TTA~\cite{wu2025mltta}, and CARVE. We use CT-CLIP and
fVLM as architecturally distinct 3D CT VLM backbones and evaluate on
multi-label abnormality diagnosis under internal and external shift, with
additional three-class and binary CT benchmarks.

The results give a consistent answer. TTA helps when the input pipeline preserves
the volumetric structure expected by the encoder and the base model is already
discriminative. In this regime, CARVE is useful because it matches the multi-label
prompt-pair structure. It estimates how many positive abnormalities should be
preserved and uses this cardinality to avoid collapsing co-occurring labels under
entropy minimization. Our contributions are summarized below.

\begin{itemize}
\itemsep0.2em
  \item We formulate unsupervised multi-label TTA for zero-shot
  3D CT VLMs, a setting not addressed by prior TTA work, and provide a diagnostic
  study showing that adaptation depends on a volumetric
  input pipeline consistent with pretraining and on transferable base
  representations rather than being automatically beneficial under shift.

  \item We identify a structural mismatch between standard entropy minimization
  and prompt-pair multi-label CT prediction. Cross-label entropy suppresses
  co-occurring abnormalities, while per-label binary entropy becomes weak near the
  decision boundary.

  \item We introduce CARVE, the first cardinality-aware and memory-efficient TTA
method for this regime, which estimates scan-specific label cardinality and
preserves multiple co-occurring abnormalities under adaptation.

\end{itemize}

\section{Related Work}
\label{sec:related}

\paragraph{3D CT vision-language models.}
Recent 3D CT VLMs learn joint representations of volumetric scans and radiology text. CT-CLIP~\cite{hamamci2024ctclip} uses a 3D ViT with global image-report contrastive pretraining on CT-RATE. fVLM~\cite{shui2025fvlm} introduces anatomy-aware fine-grained alignment, CT-GLIP~\cite{lin2024ctglip} incorporates organ grounding, and Merlin~\cite{blankemeier2026merlin} extends volumetric vision-language modeling to abdominal CT. These works focus on pretraining and static zero-shot evaluation. We instead study how such models behave under inference-time adaptation, and when adaptation is useful relative to base-model transfer.

\paragraph{Test-time adaptation.}
Tent~\cite{wang2021tent} introduced source-free TTA by minimizing prediction entropy while updating only normalization parameters. Subsequent methods improve stability and efficiency through anti-forgetting regularization~\cite{niu2022efficient}, reliable entropy minimization~\cite{niu2023towards}, and continual online adaptation~\cite{wang2022continual}. For vision-language models, TPT~\cite{shu2022tpt} adapts prompts using augmentation consistency, DiffTPT~\cite{feng2023difftpt} diversifies these augmentations with diffusion models, TDA~\cite{karmanov2024tda} replaces backpropagation with a training-free key-value cache, and RLCF~\cite{zhao2024test} drives adaptation with a CLIP reward. Recent work also revisits the evaluation assumptions behind TTA~\cite{zanella2025realistic} and adapts prediction priors at test time~\cite{zhou2025bayesian}. These approaches are primarily designed for multi-class prediction and do not address prompt-pair Bernoulli scoring or memory-constrained volumetric inference.

\paragraph{TTA for medical imaging.}
Medical TTA has been studied mostly for segmentation~\cite{yu2025medsegtta,chen2025grata}, where dense pixel-wise outputs provide rich self-supervision. For classification, prior work has largely addressed single-label prediction under label-distribution shift~\cite{ma2022ttadc}, adapting the assumed class prior rather than preserving co-occurring labels. Zero-shot multi-label CT classification is more constrained, since each scan yields only a length-$L$ prompt-pair probability vector with no caption or dense supervision to exploit, and the update direction depends on whether that vector is already separable. No prior medical TTA targets multi-label co-occurrence in this zero-shot prompt-pair setting.

\paragraph{Multi-label TTA.}
ML-TTA~\cite{wu2025mltta} studies test-time adaptation for multi-label recognition using Bound Entropy Minimization, which raises the top-$\hat{k}$ predicted labels while suppressing the rest, together with caption retrieval for pseudo-label filtering. Our setting differs in two ways. First, zero-shot 3D CT VLMs produce prompt-pair Bernoulli probabilities rather than classifier logits, so CARVE estimates the label cardinality directly from these probabilities instead of retrieving a caption. Second, volumetric inference makes dense multi-view backpropagation impractical, so CARVE scores weak 3D views without gradients and updates only on a retained subset. CARVE is designed for this setting, combining sample-specific cardinality estimation with memory-efficient retained-view adaptation.
% \mumu{
\section{Diagnosing When Adaptation Helps}
\label{sec:diagnosis}

Unsupervised adaptation does not help unconditionally, so before proposing an
objective we separate two deployment preconditions from the algorithmic problem
a method can address. We compare two architecturally distinct backbones, CT-CLIP
and fVLM, on internal CT-RATE and external RAD-ChestCT under a single episodic
protocol (\Cref{sec:setup}). First, the volumetric input must preserve the depth
structure the encoder was pretrained on, since reducing the input depth alone
substantially degrades the unadapted base before any adaptation begins
(\Cref{sec:main}). Second, the base representation must transfer, since under
severe external shift the choice of backbone moves accuracy more than any choice
of adaptation objective, a gap no CT-CLIP adapter closes (\Cref{sec:main}).
Given these preconditions, the remaining problem is that prompt-pair diagnosis
scores each abnormality as an independent Bernoulli probability rather than a
shared softmax, so cross-label entropy collapses co-occurring positives and
per-label binary entropy weakens near $p{=}0.5$ (\Cref{fig:graddiag}). CARVE
targets this mismatch by estimating a scan-level positive-label cardinality and
preserving multiple co-occurring abnormalities during adaptation.

\section{Method}
\label{sec:method}

\begin{figure*}[t]
    \centering
    \includegraphics[width=\linewidth]{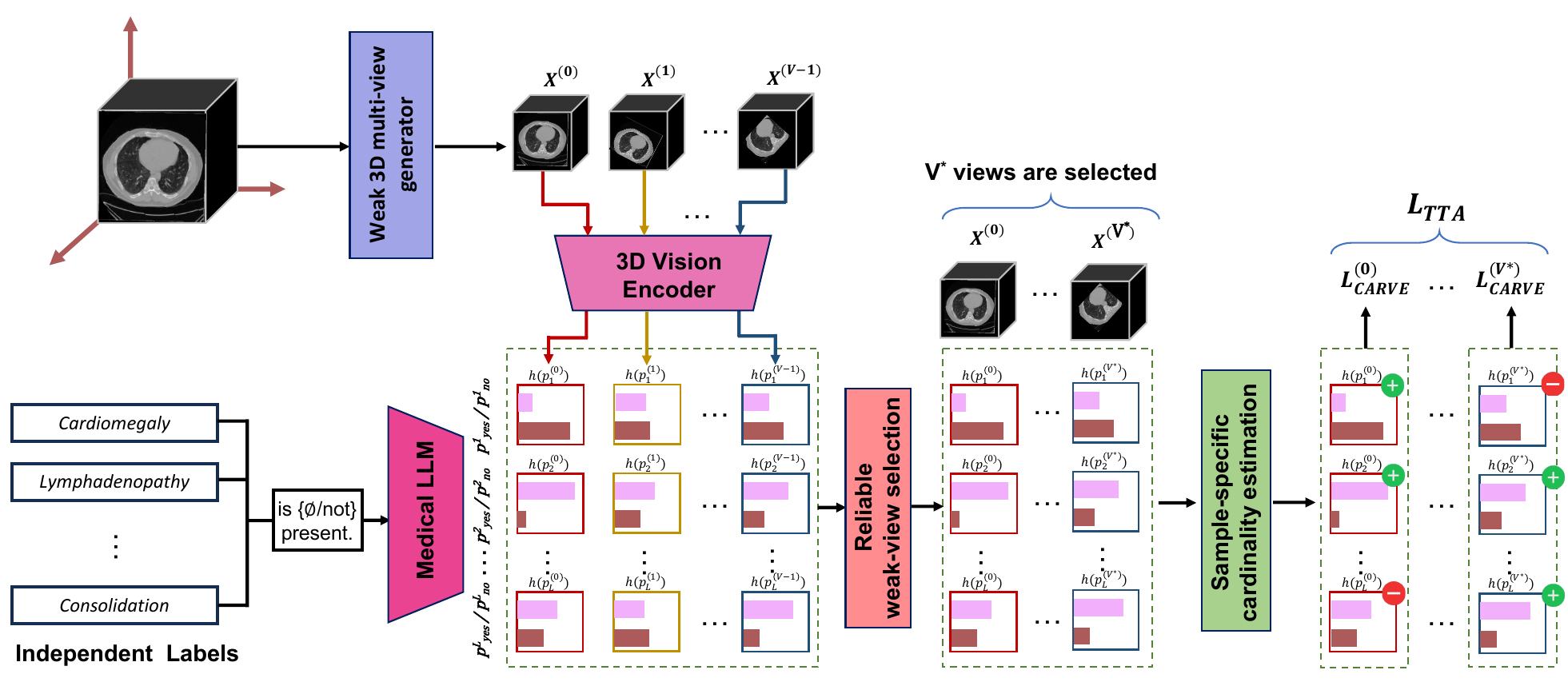}
     \caption{\textbf{CARVE overview.} Given a base 3D CT VLM (a Medical Text Encoder paired
with a 3D vision encoder), each abnormality is scored with positive/negative
prompts, yielding independent per-label probabilities. CARVE generates weak
volumetric views, retains a reliable low-entropy subset ($V^*$ of $V$ views),
estimates a sample-specific positive-label cardinality $\khat$, and updates only
the visual normalization parameters via a cardinality-aware top-$\khat$ objective
($L_{\text{CARVE}}$ per view, summed into $L_{\text{TTA}}$). Gradients are
recomputed only for the retained views, keeping volumetric test-time adaptation
memory efficient.}
    \label{fig:overview}
\end{figure*}

We present \textbf{CARVE}, a cardinality-aware, memory-safe multi-view TTA
method for zero-shot multi-label CT classification. CARVE adapts a
pretrained 3D medical VLM at inference time by updating only a restricted
subset of visual normalization parameters and replacing standard entropy
minimization with a multi-label top-$\khat$ objective. The method proceeds
in three steps. It first selects reliable weak 3D views, then estimates a
scan-specific positive-label cardinality, and finally optimizes a
cardinality-aware objective that preserves multiple co-occurring
abnormalities instead of collapsing toward a single label.

% \subsection{Zero-shot multi-label CT classification}
% \subsection{Prompt-pair zero-shot prediction}

% Let $x\!\in\!\mathbb{R}^{D\times H\times W}$ be a CT volume and
% $\{1,\dots,L\}$ denote the set of independent labels. For label $j$ we build a positive prompt
% $t_j^+$ and negative prompt $t_j^-$; the VLM yields similarity scores
% $s_j^\pm(x;\theta)$\leon{the function needs to take $t_j$ as input as well and we should define what similarity is being used}, converted to a Bernoulli presence probability via a
% two-class softmax, equivalently a logistic of the margin
% $d_j\!=\!s_j^+\!-\!s_j^-$:
% \begin{equation}
% p_j(x;\theta)=\frac{\exp(s_j^+)}{\exp(s_j^-)+\exp(s_j^+)}=\sigma(d_j).
% \label{eq:pj}
% \end{equation}
% \leon{I don't think this is correct, as it does not give a margin. It does give probability.}
% This preserves zero-shot prediction without a classifier head and
% implies adaptation acts on independent per-label probabilities.

\subsection{Per-label probabilities from prompt pairs}
Let $x\!\in\!\mathbb{R}^{D\times H\times W}$ be a CT volume and
$\{1,\dots,L\}$ denote the set of independent labels. For each label $j$ we
build a positive prompt $t_j^+$ and a negative prompt $t_j^-$. The VLM scores
each image-text pair through its similarity function $S_\theta$,
\begin{equation}
s_j^\pm(x;\theta)=S_\theta\!\big(x,\,t_j^\pm\big),
\qquad
S_\theta(x,t)=\frac{\langle f_\theta(x),\,g_\theta(t)\rangle}
{\lVert f_\theta(x)\rVert\,\lVert g_\theta(t)\rVert},
\end{equation}
the cosine similarity between the image embedding $f_\theta(x)$ and the text
embedding $g_\theta(t)$ (any backbone-specific pooling or anatomy-aware
aggregation is absorbed into $f_\theta$). The positive-negative score margin
for label $j$ is
\begin{equation}
d_j(x;\theta)=s_j^+(x;\theta)-s_j^-(x;\theta).
\end{equation}
A two-class softmax over the positive and negative prompt scores converts this
margin into a Bernoulli presence probability; equivalently, $p_j$ is the
logistic of the margin,
\begin{equation}
p_j(x;\theta)
=\frac{\exp(s_j^+)}{\exp(s_j^-)+\exp(s_j^+)}
=\sigma\!\big(d_j(x;\theta)\big).
\label{eq:pj}
\end{equation}
This preserves zero-shot prediction without a classifier head and implies
adaptation acts on independent per-label Bernoulli probabilities.

\subsection{Norm-only episodic adaptation}
Following Tent~\cite{wang2021tent}, we update only the affine visual
normalization parameters $\phi$ in selected transformer blocks, while the remaining visual parameters and all text-side parameters stay frozen; we write $\theta=(\theta_{\mathrm{frozen}},\phi)$. Adaptation is episodic, meaning that for each test scan, $\phi$ is reset to its source value, updated for a small number of gradient steps using an unsupervised objective
\begin{equation}
\phi^\star(x)=\arg\min_\phi
\mathcal{L}_{\mathrm{TTA}}(x;\theta_{\mathrm{frozen}},\phi),
\end{equation}
and then used to predict the unperturbed scan. This avoids drift across the target stream.

\subsection{Entropy-based selection of augmented views}
We form $V$ stochastically augmented copies of the input volume, which we call
\emph{views}, $x^{(v)}=\mathcal{A}_v(x)$ for $v=0,\dots,V-1$, with
$x^{(0)}=x$ the unperturbed input ($\mathcal{A}_0=\mathrm{id}$). A view here is
an augmented copy of the volume used to probe prediction stability; this is
distinct from an anatomical CT viewing plane (axial, coronal, sagittal). Each transform $\mathcal{A}_v$ applies mild volumetric distortions, using additive Gaussian intensity jitter and subtle spatial translations to preserve semantic content while introducing mild variability across views.

For each view, the model returns the per-label probability vector:
\begin{equation}
\mathbf{p}^{(v)}=\mathbf{p}(x^{(v)};\theta)
=\big(p_1^{(v)},\dots,p_L^{(v)}\big)\in[0,1]^L,
\end{equation}
where $p_j^{(v)}$ is the Bernoulli presence probability of label $j$ from
Eq.~\eqref{eq:pj}. We score view uncertainty by the mean binary entropy across
labels,
\begin{equation}
E^{(v)}=\frac{1}{L}\sum_{j=1}^{L} h\!\big(p_j^{(v)}\big),
\end{equation}
where $h(p)=-p\log p-(1-p)\log(1-p)$ is the Bernoulli entropy.
We retain the $K$ lowest-entropy views $\mathcal{V}^\star$, with
$|\mathcal{V}^\star|=K=\max\!\big(1,\mathrm{round}(V\rho)\big)$ and
$\rho\in(0,1]$ a view-retention ratio controlling the fraction of weak views
used for adaptation, and average their predictions:
\begin{equation}
\bar{\mathbf{p}}=\frac{1}{K}\sum_{v\in\mathcal{V}^\star}\mathbf{p}^{(v)}.
\end{equation}

\subsection{Sample-specific cardinality estimation} 
\label{sec:multilabel}

To resolve the objective mismatch identified in \Cref{sec:diagnosis}, CARVE estimates a scan-specific cardinality $\khat$ and uses it to define the positive set in the adaptation objective.

\noindent
\textbf{Estimating scan-specific cardinality.}
Because prompt-pair scores define independent per-label probabilities, the Bernoulli count
$\sum_{j=1}^{L}\bar{p}_j$ provides a label-free estimate of how many
abnormalities are present in the scan. We convert this soft count into a
scan-specific cardinality,
\begin{equation}
\khat=
\mathrm{clip}\!\left(
\mathrm{round}\!\left(\sum_{j=1}^{L}\bar{p}_j\right),
k_{\min},k_{\max}
\right),
\label{eq:khat}
\end{equation}
where $k_{\min}$ and $k_{\max}$ are fixed lower and upper bounds on the allowed positive-label count, with their values reported in
Appendix~\ref{app:implementation}. The lower bound preserves at least one candidate abnormality, while the upper bound prevents noisy or poorly calibrated probabilities from selecting an implausibly large positive set.
The resulting $\khat$ determines how many labels are preserved by the top-$\khat$ adaptation objective.

\subsection{Cardinality-aware retained-view objective}
Motivated by cardinality-based multi-label TTA~\cite{wu2025mltta}, CARVE
uses the scan-specific estimate $\khat$ to decide how many abnormalities
should be preserved during adaptation. For each retained view
$v\in\mathcal{V}^\star$, let
$\mathcal{P}^{(v)}=\mathrm{TopK}_{\khat}(\mathbf{p}^{(v)})$ be the index
set of the $\khat$ largest probabilities, and let $\mathcal{N}^{(v)}$ be
its complement. CARVE treats $\mathcal{P}^{(v)}$ as the scan-specific set
of likely positive labels and $\mathcal{N}^{(v)}$ as the remaining labels
to suppress:
\begin{equation}
\mathcal{L}_{\mathrm{CARVE}}^{(v)}\!=\!
\!-\!\frac{1}{|\mathcal{P}^{(v)}|}
\!\sum_{j\in\mathcal{P}^{(v)}}\! \log p_j^{(v)}
\!-\!\lambda_{\mathrm{neg}}
\frac{1}{|\mathcal{N}^{(v)}|}
\!\sum_{j\in\mathcal{N}^{(v)}}\! \log\!(1-p_j^{(v)}\!).
\label{eq:carve}
\end{equation}
The first term sharpens the estimated positive set, while the second term
discourages diffuse activation over the remaining labels. Unlike standard
entropy minimization, this objective does not force the scan toward a
single dominant abnormality and does not treat every label independently
of the scan-level cardinality.

The final adaptation loss averages over the retained views is defined as:
\begin{equation}
\mathcal{L}_{\mathrm{TTA}}=
\frac{1}{K}
\sum_{v\in\mathcal{V}^\star}
\mathcal{L}_{\mathrm{CARVE}}^{(v)}.
\label{eq:tta}
\end{equation}
We minimize it over the adapted normalization parameters $\phi$ and
then predict on the original unperturbed volume.

\subsection{Memory-efficient retained-view adaptation}
\label{sec:gatedef}

Back propagating through all augmented 3D views is impractical for
volumetric inference. CARVE therefore uses a two-stage update. It first
scores all weak views without gradients to obtain the view uncertainties
$E^{(v)}$, the retained-view average $\bar{\mathbf{p}}$, and the
cardinality estimate $\khat$. It then recomputes only the retained subset
$\mathcal{V}^\star$ with gradients for the adaptation step. This avoids
storing gradient activations for all $V$ views and limits backpropagation
to the retained subset of size $K$. CARVE also uses prediction dispersion to avoid updates when the label
probabilities carry little contrast to sharpen. We define
\begin{equation}
\sigma_{\mathrm{pred}}(x)=\mathrm{std}(\{p_1,\dots,p_L\})
\end{equation}
as a label-free gating signal. The gated variant applies the adaptation
pipeline only when $\sigma_{\mathrm{pred}}(x)>\tau$; otherwise, it returns
the unchanged zero-shot prediction. This gate is conservative. Low
dispersion indicates limited label contrast, often because probabilities
have collapsed near the decision boundary, where the unsupervised update
direction is unreliable. We use $\sigma_{\mathrm{pred}}$ only as an
abstention signal, not as a detector of scans for which adaptation is
guaranteed to help (Sec.~\ref{sec:ablation-mechanism}).

\definecolor{ctrateshade}{RGB}{226,239,218}

\begin{table*}[t]
\centering
\scriptsize
\setlength{\tabcolsep}{2pt}
\renewcommand{\arraystretch}{1.08}
\caption{ \textbf{Test-time adaptation of CT-CLIP on internal CT-RATE, across three model variants and two volumetric depths.}
}
\label{tab:main-ctclip-ctrate-internal}
\resizebox{\textwidth}{!}{%
\begin{tabular}{lcccccccccccc}
\toprule
& \multicolumn{6}{c}{$z{=}240$} & \multicolumn{6}{c}{$z{=}40$} \\
\cmidrule(lr){2-7} \cmidrule(lr){8-13}
\textbf{Method} & AUROC & AUPRC & F1 & Prec. & Rec. & Acc. & AUROC & AUPRC & F1 & Prec. & Rec. & Acc. \\
\midrule
\multicolumn{13}{l}{\cellcolor{ctrateshade}\textbf{CT-CLIP $\,\vert\,$ CT-RATE (18-abnormality multi-label)}}\\
\midrule
\multicolumn{13}{l}{\itshape Zero-shot}\\[1pt]
\quad No-TTA & $\underline{0.713}_{\pm0.0001}$ & $\underline{0.369}_{\pm0.0001}$ & $\underline{0.393}_{\pm0.0001}$ & $0.296_{\pm0.0001}$ & $0.698_{\pm0.0001}$ & $0.633_{\pm0.0001}$ & $0.588_{\pm0.0001}$ & $0.260_{\pm0.0001}$ & $\underline{0.308}_{\pm0.0001}$ & $\underline{0.198}_{\pm0.0001}$ & $0.914_{\pm0.0001}$ & $\mathbf{0.257}_{\pm0.0001}$ \\
\quad TENT \venue{ICLR 2021} & $0.712_{\pm0.0001}$ & $0.365_{\pm0.002}$ & $0.382_{\pm0.0001}$ & $0.283_{\pm0.001}$ & $\underline{0.712}_{\pm0.003}$ & $0.605_{\pm0.003}$ & $\underline{0.591}_{\pm0.004}$ & $\underline{0.266}_{\pm0.001}$ & $\underline{0.308}_{\pm0.001}$ & $0.196_{\pm0.0001}$ & $\underline{0.936}_{\pm0.003}$ & $0.237_{\pm0.001}$ \\
\quad RLCF \venue{ICLR 2024} & $\underline{0.713}_{\pm0.0001}$ & $0.366_{\pm0.0001}$ & $0.391_{\pm0.0001}$ & $\underline{0.297}_{\pm0.0001}$ & $0.682_{\pm0.0001}$ & $\underline{0.636}_{\pm0.0001}$ & $0.590_{\pm0.0001}$ & $0.263_{\pm0.0001}$ & $\underline{0.308}_{\pm0.0001}$ & $0.197_{\pm0.0001}$ & $0.925_{\pm0.0001}$ & $0.247_{\pm0.0001}$ \\
\quad ML-TTA \venue{ICLR 2025} & $0.712_{\pm0.0001}$ & $0.365_{\pm0.002}$ & $0.382_{\pm0.0001}$ & $0.283_{\pm0.001}$ & $\underline{0.712}_{\pm0.003}$ & $0.605_{\pm0.003}$ & $\underline{0.591}_{\pm0.004}$ & $\underline{0.266}_{\pm0.001}$ & $\underline{0.308}_{\pm0.001}$ & $0.196_{\pm0.0001}$ & $\underline{0.936}_{\pm0.003}$ & $0.237_{\pm0.0001}$ \\
\rowcolor{oursgray} \quad \textbf{CARVE} & $\mathbf{0.749}_{\pm0.0001}$ & $\mathbf{0.386}_{\pm0.003}$ & $\mathbf{0.406}_{\pm0.0001}$ & $\mathbf{0.303}_{\pm0.001}$ & $\mathbf{0.747}_{\pm0.003}$ & $\mathbf{0.647}_{\pm0.003}$ & $\mathbf{0.621}_{\pm0.004}$ & $\mathbf{0.279}_{\pm0.001}$ & $\mathbf{0.324}_{\pm0.001}$ & $\mathbf{0.206}_{\pm0.0001}$ & $\mathbf{0.983}_{\pm0.003}$ & $\underline{0.249}_{\pm0.001}$ \\
\addlinespace[2pt]

\multicolumn{13}{l}{\itshape Vocab-FT}\\[1pt]
\quad No-TTA & $0.763_{\pm0.0001}$ & $0.442_{\pm0.0001}$ & $\mathbf{0.262}_{\pm0.0001}$ & $\mathbf{0.584}_{\pm0.0001}$ & $\mathbf{0.208}_{\pm0.0001}$ & $0.822_{\pm0.0001}$ & $0.654_{\pm0.0001}$ & $0.315_{\pm0.0001}$ & $\mathbf{0.250}_{\pm0.0001}$ & $0.387_{\pm0.0001}$ & $\mathbf{0.279}_{\pm0.0001}$ & $0.762_{\pm0.0001}$ \\
\quad TENT \venue{ICLR 2021} & $0.765_{\pm0.0001}$ & $\underline{0.446}_{\pm0.001}$ & $0.250_{\pm0.001}$ & $0.516_{\pm0.002}$ & $0.197_{\pm0.0001}$ & $\underline{0.823}_{\pm0.0001}$ & $\underline{0.661}_{\pm0.0001}$ & $\underline{0.319}_{\pm0.001}$ & $0.238_{\pm0.003}$ & $\underline{0.396}_{\pm0.029}$ & $0.265_{\pm0.001}$ & $\underline{0.767}_{\pm0.001}$ \\
\quad RLCF \venue{ICLR 2024} & $0.764_{\pm0.0001}$ & $0.444_{\pm0.0001}$ & $0.247_{\pm0.0001}$ & $0.512_{\pm0.0001}$ & $0.194_{\pm0.0001}$ & $\underline{0.823}_{\pm0.0001}$ & $0.656_{\pm0.0001}$ & $0.316_{\pm0.0001}$ & $\underline{0.241}_{\pm0.0001}$ & $0.386_{\pm0.0001}$ & $0.266_{\pm0.0001}$ & $0.764_{\pm0.0001}$ \\
\quad ML-TTA \venue{ICLR 2025} & $\underline{0.766}_{\pm0.0001}$ & $\underline{0.446}_{\pm0.001}$ & $0.250_{\pm0.001}$ & $0.516_{\pm0.002}$ & $0.197_{\pm0.0001}$ & $\underline{0.823}_{\pm0.0001}$ & $\underline{0.661}_{\pm0.0001}$ & $\underline{0.319}_{\pm0.001}$ & $0.238_{\pm0.003}$ & $\underline{0.396}_{\pm0.029}$ & $0.264_{\pm0.001}$ & $\underline{0.767}_{\pm0.0001}$ \\
\rowcolor{oursgray} \quad \textbf{CARVE} & $\mathbf{0.804}_{\pm0.0001}$ & $\mathbf{0.468}_{\pm0.001}$ & $\underline{0.261}_{\pm0.001}$ & $\underline{0.542}_{\pm0.002}$ & $\underline{0.207}_{\pm0.0001}$ & $\mathbf{0.864}_{\pm0.0001}$ & $\mathbf{0.694}_{\pm0.0001}$ & $\mathbf{0.335}_{\pm0.001}$ & $\mathbf{0.250}_{\pm0.003}$ & $\mathbf{0.417}_{\pm0.032}$ & $\underline{0.278}_{\pm0.001}$ & $\mathbf{0.806}_{\pm0.001}$ \\
\addlinespace[2pt]

\multicolumn{13}{l}{\itshape Class-FT}\\[1pt]
\quad No-TTA & $0.520_{\pm0.009}$ & $0.218_{\pm0.004}$ & $\underline{0.271}_{\pm0.012}$ & $0.205_{\pm0.006}$ & $0.500_{\pm0.092}$ & $0.521_{\pm0.064}$ & $\underline{0.506}_{\pm0.005}$ & $\underline{0.218}_{\pm0.005}$ & $\underline{0.260}_{\pm0.001}$ & $0.198_{\pm0.003}$ & $\underline{0.447}_{\pm0.024}$ & $0.537_{\pm0.015}$ \\
\quad TENT \venue{ICLR 2021} & $\underline{0.524}_{\pm0.012}$ & $0.217_{\pm0.007}$ & $0.232_{\pm0.084}$ & $0.188_{\pm0.020}$ & $0.510_{\pm0.283}$ & $0.512_{\pm0.175}$ & $0.488_{\pm0.002}$ & $0.201_{\pm0.004}$ & $0.239_{\pm0.033}$ & $0.196_{\pm0.003}$ & $0.425_{\pm0.172}$ & $0.538_{\pm0.117}$ \\
\quad RLCF \venue{ICLR 2024} & $\underline{0.524}_{\pm0.006}$ & $\underline{0.219}_{\pm0.002}$ & $0.247_{\pm0.050}$ & $\underline{0.208}_{\pm0.005}$ & $0.464_{\pm0.189}$ & $\underline{0.551}_{\pm0.115}$ & $\underline{0.506}_{\pm0.004}$ & $0.214_{\pm0.002}$ & $0.253_{\pm0.001}$ & $\underline{0.203}_{\pm0.016}$ & $0.422_{\pm0.077}$ & $\underline{0.555}_{\pm0.073}$ \\
\quad ML-TTA \venue{ICLR 2025} & $\underline{0.524}_{\pm0.012}$ & $0.217_{\pm0.007}$ & $0.232_{\pm0.085}$ & $0.188_{\pm0.020}$ & $\underline{0.511}_{\pm0.284}$ & $0.513_{\pm0.175}$ & $0.488_{\pm0.003}$ & $0.201_{\pm0.005}$ & $0.239_{\pm0.035}$ & $0.195_{\pm0.002}$ & $0.424_{\pm0.175}$ & $0.538_{\pm0.116}$ \\
\rowcolor{oursgray} \quad \textbf{CARVE} & $\mathbf{0.556}_{\pm0.006}$ & $\mathbf{0.233}_{\pm0.001}$ & $\mathbf{0.291}_{\pm0.022}$ & $\mathbf{0.216}_{\pm0.005}$ & $\mathbf{0.602}_{\pm0.206}$ & $\mathbf{0.584}_{\pm0.119}$ & $\mathbf{0.531}_{\pm0.005}$ & $\mathbf{0.228}_{\pm0.006}$ & $\mathbf{0.274}_{\pm0.002}$ & $\mathbf{0.208}_{\pm0.003}$ & $\mathbf{0.511}_{\pm0.085}$ & $\mathbf{0.602}_{\pm0.070}$ \\
\bottomrule
\end{tabular}%
}
\end{table*}

\begin{table*}[t]
\centering
\scriptsize
\setlength{\tabcolsep}{2pt}
\renewcommand{\arraystretch}{1.08}
\caption{ \textbf{Test-time adaptation of CT-CLIP on external RAD-ChestCT, across three model variants and two volumetric depths.}
}
\label{tab:main-ctclip-rad-external}
\resizebox{\textwidth}{!}{%
\begin{tabular}{lcccccccccccc}
\toprule
& \multicolumn{6}{c}{$z{=}240$} & \multicolumn{6}{c}{$z{=}40$} \\
\cmidrule(lr){2-7} \cmidrule(lr){8-13}
\textbf{Method} & AUROC & AUPRC & F1 & Prec. & Rec. & Acc. & AUROC & AUPRC & F1 & Prec. & Rec. & Acc. \\
\midrule
\multicolumn{13}{l}{\cellcolor{headerblue}\textbf{CT-CLIP $\,\vert\,$ RAD-ChestCT (16-abnormality multi-label)}}\\
\midrule
\multicolumn{13}{l}{\itshape Zero-shot}\\[1pt]
\quad No-TTA & $0.499_{\pm0.0001}$ & $0.283_{\pm0.0001}$ & $0.298_{\pm0.0001}$ & $0.282_{\pm0.0001}$ & $0.671_{\pm0.0001}$ & $\mathbf{0.373}_{\pm0.0001}$ & $0.511_{\pm0.0001}$ & $0.297_{\pm0.0001}$ & $0.249_{\pm0.0001}$ & $\underline{0.283}_{\pm0.0001}$ & $0.400_{\pm0.0001}$ & $\mathbf{0.524}_{\pm0.0001}$ \\
\quad TENT \venue{ICLR 2021} & $\underline{0.508}_{\pm0.002}$ & $\underline{0.290}_{\pm0.0001}$ & $0.304_{\pm0.003}$ & $0.278_{\pm0.006}$ & $0.706_{\pm0.003}$ & $0.356_{\pm0.001}$ & $\underline{0.514}_{\pm0.001}$ & $0.297_{\pm0.002}$ & $0.289_{\pm0.003}$ & $\underline{0.283}_{\pm0.0001}$ & $0.498_{\pm0.007}$ & $0.484_{\pm0.002}$ \\
\quad RLCF \venue{ICLR 2024} & $0.504_{\pm0.0001}$ & $0.287_{\pm0.0001}$ & $\mathbf{0.322}_{\pm0.0001}$ & $\underline{0.287}_{\pm0.0001}$ & $\underline{0.735}_{\pm0.0001}$ & $\underline{0.361}_{\pm0.0001}$ & $0.512_{\pm0.0001}$ & $\underline{0.300}_{\pm0.0001}$ & $\underline{0.293}_{\pm0.0001}$ & $0.281_{\pm0.0001}$ & $\underline{0.502}_{\pm0.0001}$ & $0.489_{\pm0.0001}$ \\
\quad ML-TTA \venue{ICLR 2025} & $\underline{0.508}_{\pm0.002}$ & $\underline{0.290}_{\pm0.0001}$ & $0.303_{\pm0.002}$ & $0.278_{\pm0.005}$ & $0.706_{\pm0.002}$ & $0.355_{\pm0.001}$ & $\underline{0.514}_{\pm0.001}$ & $0.297_{\pm0.002}$ & $0.289_{\pm0.003}$ & $0.282_{\pm0.0001}$ & $0.498_{\pm0.007}$ & $0.484_{\pm0.002}$ \\
\rowcolor{oursgray} \quad \textbf{CARVE} & $\mathbf{0.533}_{\pm0.003}$ & $\mathbf{0.304}_{\pm0.0001}$ & $\underline{0.319}_{\pm0.003}$ & $\mathbf{0.292}_{\pm0.006}$ & $\mathbf{0.741}_{\pm0.003}$ & $\mathbf{0.373}_{\pm0.001}$ & $\mathbf{0.540}_{\pm0.001}$ & $\mathbf{0.312}_{\pm0.002}$ & $\mathbf{0.304}_{\pm0.003}$ & $\mathbf{0.298}_{\pm0.001}$ & $\mathbf{0.523}_{\pm0.007}$ & $\underline{0.512}_{\pm0.001}$ \\
\addlinespace[2pt]
\multicolumn{13}{l}{\itshape Vocab-FT}\\[1pt]
\quad No-TTA & $\underline{0.496}_{\pm0.0001}$ & $\underline{0.285}_{\pm0.0001}$ & $\mathbf{0.054}_{\pm0.0001}$ & $0.312_{\pm0.0001}$ & $\mathbf{0.039}_{\pm0.0001}$ & $\underline{0.727}_{\pm0.0001}$ & $\underline{0.471}_{\pm0.0001}$ & $\underline{0.276}_{\pm0.0001}$ & $\mathbf{0.059}_{\pm0.0001}$ & $0.243_{\pm0.0001}$ & $\mathbf{0.053}_{\pm0.0001}$ & $0.722_{\pm0.0001}$ \\
\quad TENT \venue{ICLR 2021} & $0.493_{\pm0.001}$ & $0.283_{\pm0.0001}$ & $0.048_{\pm0.0001}$ & $\underline{0.403}_{\pm0.005}$ & $0.034_{\pm0.0001}$ & $\underline{0.727}_{\pm0.0001}$ & $0.467_{\pm0.001}$ & $\underline{0.276}_{\pm0.0001}$ & $\underline{0.056}_{\pm0.001}$ & $0.313_{\pm0.018}$ & $0.049_{\pm0.0001}$ & $\underline{0.723}_{\pm0.0001}$ \\
\quad RLCF \venue{ICLR 2024} & $\underline{0.496}_{\pm0.0001}$ & $0.284_{\pm0.0001}$ & $0.039_{\pm0.0001}$ & $0.365_{\pm0.0001}$ & $0.026_{\pm0.0001}$ & $\underline{0.727}_{\pm0.0001}$ & $0.464_{\pm0.0001}$ & $0.273_{\pm0.0001}$ & $0.051_{\pm0.0001}$ & $\underline{0.317}_{\pm0.0001}$ & $0.043_{\pm0.0001}$ & $\underline{0.723}_{\pm0.0001}$ \\
\quad ML-TTA \venue{ICLR 2025} & $0.493_{\pm0.001}$ & $0.283_{\pm0.0001}$ & $0.048_{\pm0.0001}$ & $\underline{0.403}_{\pm0.006}$ & $0.034_{\pm0.0001}$ & $\underline{0.727}_{\pm0.0001}$ & $0.467_{\pm0.001}$ & $\underline{0.276}_{\pm0.0001}$ & $\underline{0.056}_{\pm0.001}$ & $0.313_{\pm0.018}$ & $0.049_{\pm0.0001}$ & $\underline{0.723}_{\pm0.0001}$ \\
\rowcolor{oursgray} \quad \textbf{CARVE} & $\mathbf{0.518}_{\pm0.001}$ & $\mathbf{0.297}_{\pm0.0001}$ & $\underline{0.050}_{\pm0.0001}$ & $\mathbf{0.424}_{\pm0.006}$ & $\underline{0.036}_{\pm0.0001}$ & $\mathbf{0.763}_{\pm0.0001}$ & $\mathbf{0.490}_{\pm0.001}$ & $\mathbf{0.290}_{\pm0.001}$ & $\mathbf{0.059}_{\pm0.001}$ & $\mathbf{0.329}_{\pm0.019}$ & $\underline{0.051}_{\pm0.0001}$ & $\mathbf{0.759}_{\pm0.0001}$ \\
\addlinespace[2pt]
\multicolumn{13}{l}{\itshape Class-FT}\\[1pt]
\quad No-TTA & $\underline{0.523}_{\pm0.014}$ & $\underline{0.303}_{\pm0.002}$ & $0.290_{\pm0.009}$ & $0.282_{\pm0.011}$ & $0.492_{\pm0.003}$ & $0.488_{\pm0.025}$ & $0.509_{\pm0.005}$ & $0.287_{\pm0.004}$ & $0.305_{\pm0.014}$ & $0.278_{\pm0.001}$ & $0.453_{\pm0.066}$ & $0.537_{\pm0.035}$ \\
\quad TENT \venue{ICLR 2021} & $0.512_{\pm0.004}$ & $0.290_{\pm0.002}$ & $\underline{0.301}_{\pm0.048}$ & $\underline{0.287}_{\pm0.013}$ & $\underline{0.560}_{\pm0.119}$ & $0.476_{\pm0.061}$ & $0.508_{\pm0.015}$ & $0.289_{\pm0.004}$ & $0.323_{\pm0.030}$ & $\underline{0.283}_{\pm0.009}$ & $0.542_{\pm0.068}$ & $0.512_{\pm0.001}$ \\
\quad RLCF \venue{ICLR 2024} & $0.509_{\pm0.021}$ & $0.296_{\pm0.001}$ & $0.218_{\pm0.011}$ & $0.260_{\pm0.027}$ & $0.302_{\pm0.018}$ & $\mathbf{0.593}_{\pm0.024}$ & $\underline{0.518}_{\pm0.008}$ & $\underline{0.293}_{\pm0.003}$ & $0.293_{\pm0.003}$ & $0.278_{\pm0.001}$ & $0.453_{\pm0.068}$ & $\underline{0.539}_{\pm0.059}$ \\
\quad ML-TTA \venue{ICLR 2025} & $0.512_{\pm0.004}$ & $0.290_{\pm0.002}$ & $\underline{0.301}_{\pm0.048}$ & $\underline{0.287}_{\pm0.013}$ & $0.559_{\pm0.120}$ & $0.476_{\pm0.061}$ & $0.508_{\pm0.015}$ & $0.290_{\pm0.004}$ & $\underline{0.324}_{\pm0.029}$ & $\underline{0.283}_{\pm0.008}$ & $\underline{0.543}_{\pm0.068}$ & $0.512_{\pm0.001}$ \\
\rowcolor{oursgray} \quad \textbf{CARVE} & $\mathbf{0.550}_{\pm0.014}$ & $\mathbf{0.318}_{\pm0.002}$ & $\mathbf{0.325}_{\pm0.038}$ & $\mathbf{0.307}_{\pm0.004}$ & $\mathbf{0.597}_{\pm0.112}$ & $\underline{0.538}_{\pm0.009}$ & $\mathbf{0.538}_{\pm0.010}$ & $\mathbf{0.306}_{\pm0.002}$ & $\mathbf{0.347}_{\pm0.023}$ & $\mathbf{0.298}_{\pm0.010}$ & $\mathbf{0.571}_{\pm0.072}$ & $\mathbf{0.564}_{\pm0.037}$ \\
\bottomrule
\end{tabular}%
}
\end{table*}

\begin{table*}[t]
\centering
\scriptsize
\setlength{\tabcolsep}{2.8pt}
\renewcommand{\arraystretch}{1.06}
\caption{\textbf{Test-time adaptation of fVLM on  RAD-ChestCT and CT-RATE at native depth $z{=}240$.}}
\label{tab:main-fvlm-sidebyside}

\definecolor{radshade}{RGB}{221,235,247}
\definecolor{ctrateshade}{RGB}{226,239,218}
\definecolor{oursgray}{RGB}{235,235,235}

\resizebox{\textwidth}{!}{%
\begin{tabular}{lcccccc|cccccc}
\toprule
& \multicolumn{6}{c|}{\cellcolor{radshade}\textbf{RAD-ChestCT}} 
& \multicolumn{6}{c}{\cellcolor{ctrateshade}\textbf{CT-RATE}} \\
\cmidrule(r){2-7} \cmidrule(l){8-13}
\textbf{Method} 
& \textbf{AUROC} & \textbf{AUPRC} & \textbf{F1} & \textbf{Prec.} & \textbf{Rec.} & \textbf{Acc.}
& \textbf{AUROC} & \textbf{AUPRC} & \textbf{F1} & \textbf{Prec.} & \textbf{Rec.} & \textbf{Acc.} \\
\midrule
\multicolumn{13}{l}{\itshape Zero-shot}\\[1pt]

\quad No-TTA
& $\underline{0.612}_{\pm0.0001}$ & $\underline{0.382}_{\pm0.0001}$ & $\underline{0.395}_{\pm0.0001}$ & $\underline{0.299}_{\pm0.0001}$ & $\underline{0.814}_{\pm0.0001}$ & $\underline{0.422}_{\pm0.0001}$
& $\underline{0.687}_{\pm0.0001}$ & $\underline{0.329}_{\pm0.0001}$ & $\underline{0.376}_{\pm0.0001}$ & $\underline{0.278}_{\pm0.0001}$ & $\underline{0.718}_{\pm0.0001}$ & $\underline{0.589}_{\pm0.0001}$ \\

\quad TENT \venue{ICLR 2021}
& $\underline{0.612}_{\pm0.0001}$ & $\underline{0.382}_{\pm0.0001}$ & $\underline{0.395}_{\pm0.0001}$ & $\underline{0.299}_{\pm0.0001}$ & $\underline{0.814}_{\pm0.0001}$ & $\underline{0.422}_{\pm0.0001}$
& $\underline{0.687}_{\pm0.0001}$ & $\underline{0.329}_{\pm0.0001}$ & $\underline{0.376}_{\pm0.0001}$ & $0.277_{\pm0.0001}$ & $\underline{0.718}_{\pm0.0001}$ & $0.588_{\pm0.0001}$ \\

\quad RLCF \venue{ICLR 2024}
& $\underline{0.612}_{\pm0.0001}$ & $\underline{0.382}_{\pm0.0001}$ & $\underline{0.395}_{\pm0.0001}$ & $\underline{0.299}_{\pm0.0001}$ & $\underline{0.814}_{\pm0.0001}$ & $0.421_{\pm0.0001}$
& $\underline{0.687}_{\pm0.0001}$ & $\underline{0.329}_{\pm0.0001}$ & $\underline{0.376}_{\pm0.0001}$ & $0.277_{\pm0.0001}$ & $\underline{0.718}_{\pm0.0001}$ & $0.587_{\pm0.0001}$ \\

\quad ML-TTA \venue{ICLR 2025}
& $\underline{0.612}_{\pm0.0001}$ & $\underline{0.382}_{\pm0.0001}$ & $\underline{0.395}_{\pm0.0001}$ & $\underline{0.299}_{\pm0.0001}$ & $\underline{0.814}_{\pm0.0001}$ & $\underline{0.422}_{\pm0.0001}$
& $\underline{0.687}_{\pm0.0001}$ & $\underline{0.329}_{\pm0.0001}$ & $\underline{0.376}_{\pm0.0001}$ & $0.277_{\pm0.0001}$ & $\underline{0.718}_{\pm0.0001}$ & $0.587_{\pm0.0001}$ \\

\rowcolor{oursgray}
\quad \textbf{CARVE}
& $\mathbf{0.642}_{\pm0.0001}$ & $\mathbf{0.401}_{\pm0.0001}$ & $\mathbf{0.415}_{\pm0.0001}$ & $\mathbf{0.314}_{\pm0.0001}$ & $\mathbf{0.855}_{\pm0.0001}$ & $\mathbf{0.444}_{\pm0.0001}$
& $\mathbf{0.721}_{\pm0.0001}$ & $\mathbf{0.346}_{\pm0.0001}$ & $\mathbf{0.395}_{\pm0.0001}$ & $\mathbf{0.292}_{\pm0.0001}$ & $\mathbf{0.754}_{\pm0.0001}$ & $\mathbf{0.618}_{\pm0.0001}$ \\
\bottomrule
\end{tabular}%
}
\end{table*}

\definecolor{ccshiade}{RGB}{234,229,247}

\begin{table*}[t]
\centering
\scriptsize
\setlength{\tabcolsep}{2pt}
\renewcommand{\arraystretch}{1.08}
\caption{ \textbf{Generalization of CT-CLIP test-time adaptation to the three-label CC-CCII setting.}
}
\label{tab:gen-ctclip-ccccii-3-labels}
\resizebox{\textwidth}{!}{%
\begin{tabular}{lcccccccccccc}
\toprule
& \multicolumn{6}{c}{$z{=}240$} & \multicolumn{6}{c}{$z{=}40$} \\
\cmidrule(lr){2-7} \cmidrule(lr){8-13}
\textbf{Method} & AUROC & AUPRC & F1 & Prec. & Rec. & Acc. & AUROC & AUPRC & F1 & Prec. & Rec. & Acc. \\
\midrule
\multicolumn{13}{l}{\cellcolor{ccshiade}\textbf{CT-CLIP $\,\vert\,$ CC-CCII (3-abnormality multi-label)}}\\
\midrule
\multicolumn{13}{l}{\itshape Zero-shot}\\[1pt]
\quad No-TTA & $0.511_{\pm0.0001}$ & $\underline{0.373}_{\pm0.0001}$ & $\underline{0.340}_{\pm0.0001}$ & $\underline{0.374}_{\pm0.0001}$ & $\underline{0.313}_{\pm0.0001}$ & $0.585_{\pm0.0001}$ & $0.477_{\pm0.0001}$ & $\underline{0.347}_{\pm0.0001}$ & $0.146_{\pm0.0001}$ & $0.306_{\pm0.0001}$ & $0.098_{\pm0.0001}$ & $\underline{0.635}_{\pm0.0001}$ \\
\quad TENT \venue{ICLR 2021} & $\underline{0.517}_{\pm0.001}$ & $0.364_{\pm0.001}$ & $0.230_{\pm0.006}$ & $0.337_{\pm0.011}$ & $0.178_{\pm0.005}$ & $\underline{0.605}_{\pm0.007}$ & $\underline{0.482}_{\pm0.001}$ & $\underline{0.347}_{\pm0.0001}$ & $0.215_{\pm0.002}$ & $0.307_{\pm0.003}$ & $\underline{0.167}_{\pm0.004}$ & $0.605_{\pm0.003}$ \\
\quad RLCF \venue{ICLR 2024} & $0.514_{\pm0.0001}$ & $0.370_{\pm0.0001}$ & $0.261_{\pm0.0001}$ & $0.334_{\pm0.0001}$ & $0.216_{\pm0.0001}$ & $0.580_{\pm0.0001}$ & $0.479_{\pm0.0001}$ & $\underline{0.347}_{\pm0.0001}$ & $\mathbf{0.229}_{\pm0.0001}$ & $\mathbf{0.331}_{\pm0.0001}$ & $\mathbf{0.176}_{\pm0.0001}$ & $0.612_{\pm0.0001}$ \\
\quad ML-TTA \venue{ICLR 2025} & $\underline{0.517}_{\pm0.001}$ & $0.364_{\pm0.001}$ & $0.230_{\pm0.006}$ & $0.337_{\pm0.011}$ & $0.178_{\pm0.005}$ & $\underline{0.605}_{\pm0.007}$ & $\underline{0.482}_{\pm0.001}$ & $\underline{0.347}_{\pm0.001}$ & $0.215_{\pm0.002}$ & $0.307_{\pm0.003}$ & $\underline{0.167}_{\pm0.004}$ & $0.605_{\pm0.003}$ \\
\rowcolor{oursgray} \quad \textbf{CARVE} & $\mathbf{0.543}_{\pm0.001}$ & $\mathbf{0.392}_{\pm0.0001}$ & $\mathbf{0.357}_{\pm0.0001}$ & $\mathbf{0.392}_{\pm0.0001}$ & $\mathbf{0.329}_{\pm0.0001}$ & $\mathbf{0.627}_{\pm0.018}$ & $\mathbf{0.506}_{\pm0.001}$ & $\mathbf{0.365}_{\pm0.0001}$ & $\underline{0.226}_{\pm0.002}$ & $\underline{0.323}_{\pm0.003}$ & $\mathbf{0.176}_{\pm0.005}$ & $\mathbf{0.666}_{\pm0.0001}$ \\
\addlinespace[2pt]
\multicolumn{13}{l}{\itshape Vocab-FT}\\[1pt]
\quad No-TTA & $\underline{0.501}_{\pm0.0001}$ & $0.361_{\pm0.0001}$ & $\underline{0.137}_{\pm0.0001}$ & $\underline{0.381}_{\pm0.0001}$ & $\underline{0.084}_{\pm0.0001}$ & $\underline{0.652}_{\pm0.0001}$ & $0.495_{\pm0.0001}$ & $0.383_{\pm0.0001}$ & $\underline{0.174}_{\pm0.0001}$ & $0.354_{\pm0.0001}$ & $\underline{0.116}_{\pm0.0001}$ & $0.637_{\pm0.0001}$ \\
\quad TENT \venue{ICLR 2021} & $0.500_{\pm0.0001}$ & $\underline{0.365}_{\pm0.001}$ & $0.105_{\pm0.013}$ & $0.290_{\pm0.068}$ & $0.065_{\pm0.006}$ & $0.638_{\pm0.012}$ & $\underline{0.497}_{\pm0.0001}$ & $0.384_{\pm0.0001}$ & $0.079_{\pm0.009}$ & $0.333_{\pm0.0001}$ & $0.045_{\pm0.006}$ & $\underline{0.654}_{\pm0.0001}$ \\
\quad RLCF \venue{ICLR 2024} & $0.500_{\pm0.0001}$ & $0.362_{\pm0.0001}$ & $0.090_{\pm0.0001}$ & $0.270_{\pm0.0001}$ & $0.054_{\pm0.0001}$ & $0.644_{\pm0.0001}$ & $0.496_{\pm0.0001}$ & $\underline{0.387}_{\pm0.0001}$ & $0.143_{\pm0.0001}$ & $\underline{0.367}_{\pm0.0001}$ & $0.089_{\pm0.0001}$ & $0.647_{\pm0.0001}$ \\
\quad ML-TTA \venue{ICLR 2025} & $0.500_{\pm0.0001}$ & $\underline{0.365}_{\pm0.001}$ & $0.105_{\pm0.013}$ & $0.290_{\pm0.068}$ & $0.065_{\pm0.006}$ & $0.638_{\pm0.012}$ & $\underline{0.497}_{\pm0.0001}$ & $0.384_{\pm0.0001}$ & $0.079_{\pm0.009}$ & $0.333_{\pm0.0001}$ & $0.045_{\pm0.006}$ & $\underline{0.654}_{\pm0.0001}$ \\
\rowcolor{oursgray} \quad \textbf{CARVE} & $\mathbf{0.527}_{\pm0.0001}$ & $\mathbf{0.383}_{\pm0.001}$ & $\mathbf{0.144}_{\pm0.0001}$ & $\mathbf{0.400}_{\pm0.0001}$ & $\mathbf{0.088}_{\pm0.0001}$ & $\mathbf{0.684}_{\pm0.0001}$ & $\mathbf{0.522}_{\pm0.0001}$ & $\mathbf{0.404}_{\pm0.0001}$ & $\mathbf{0.183}_{\pm0.0001}$ & $\mathbf{0.372}_{\pm0.0001}$ & $\mathbf{0.121}_{\pm0.0001}$ & $\mathbf{0.687}_{\pm0.0001}$ \\
\addlinespace[2pt]
\multicolumn{13}{l}{\itshape Class-FT}\\[1pt]
\quad No-TTA & $0.499_{\pm0.026}$ & $\underline{0.362}_{\pm0.014}$ & $\underline{0.373}_{\pm0.001}$ & $0.337_{\pm0.019}$ & $\underline{0.421}_{\pm0.027}$ & $0.532_{\pm0.033}$ & $0.507_{\pm0.002}$ & $0.359_{\pm0.004}$ & $\underline{0.389}_{\pm0.025}$ & $0.331_{\pm0.0001}$ & $\underline{0.479}_{\pm0.072}$ & $0.507_{\pm0.026}$ \\
\quad TENT \venue{ICLR 2021} & $0.501_{\pm0.003}$ & $0.361_{\pm0.012}$ & $0.293_{\pm0.170}$ & $0.373_{\pm0.060}$ & $0.336_{\pm0.315}$ & $0.560_{\pm0.122}$ & $\underline{0.515}_{\pm0.024}$ & $0.361_{\pm0.012}$ & $0.335_{\pm0.186}$ & $\underline{0.343}_{\pm0.022}$ & $0.413_{\pm0.375}$ & $\underline{0.553}_{\pm0.087}$ \\
\quad RLCF \venue{ICLR 2024} & $\underline{0.505}_{\pm0.007}$ & $0.356_{\pm0.003}$ & $0.285_{\pm0.125}$ & $0.359_{\pm0.029}$ & $0.277_{\pm0.205}$ & $\underline{0.589}_{\pm0.072}$ & $0.508_{\pm0.010}$ & $0.360_{\pm0.011}$ & $0.355_{\pm0.057}$ & $0.336_{\pm0.011}$ & $0.397_{\pm0.148}$ & $0.538_{\pm0.059}$ \\
\quad ML-TTA \venue{ICLR 2025} & $0.501_{\pm0.003}$ & $0.361_{\pm0.012}$ & $0.293_{\pm0.169}$ & $\underline{0.385}_{\pm0.077}$ & $0.336_{\pm0.315}$ & $0.562_{\pm0.124}$ & $\underline{0.515}_{\pm0.024}$ & $\underline{0.362}_{\pm0.011}$ & $0.335_{\pm0.186}$ & $\underline{0.343}_{\pm0.022}$ & $0.413_{\pm0.375}$ & $\underline{0.553}_{\pm0.087}$ \\
\rowcolor{oursgray} \quad \textbf{CARVE} & $\mathbf{0.534}_{\pm0.015}$ & $\mathbf{0.381}_{\pm0.014}$ & $\mathbf{0.413}_{\pm0.029}$ & $\mathbf{0.392}_{\pm0.063}$ & $\mathbf{0.504}_{\pm0.117}$ & $\mathbf{0.607}_{\pm0.103}$ & $\mathbf{0.544}_{\pm0.019}$ & $\mathbf{0.384}_{\pm0.006}$ & $\mathbf{0.440}_{\pm0.070}$ & $\mathbf{0.366}_{\pm0.013}$ & $\mathbf{0.581}_{\pm0.187}$ & $\mathbf{0.582}_{\pm0.097}$ \\
\bottomrule
\end{tabular}%
}
\end{table*}

\definecolor{lunashade}{RGB}{250,245,210}
\begin{table*}[t]
\centering
\scriptsize
\setlength{\tabcolsep}{2pt}
\renewcommand{\arraystretch}{1.08}
\caption{\textbf{Generalization of CT-CLIP test-time adaptation to the binary-class LUNA16 setting.}}
\label{tab:gen-ctclip-luna16-2-labels}
\resizebox{\textwidth}{!}{%
\begin{tabular}{lcccccccccccc}
\toprule
& \multicolumn{6}{c}{$z{=}240$} & \multicolumn{6}{c}{$z{=}40$} \\
\cmidrule(lr){2-7} \cmidrule(lr){8-13}
\textbf{Method} & AUROC & AUPRC & F1 & Prec. & Rec. & Acc. & AUROC & AUPRC & F1 & Prec. & Rec. & Acc. \\
\midrule
\multicolumn{13}{l}{\cellcolor{lunashade}\textbf{CT-CLIP $\,\vert\,$ LUNA16 (2 labels)}}\\
\midrule
\multicolumn{13}{l}{\itshape Zero-shot}\\[1pt]
\quad No-TTA & $\underline{0.527}_{\pm0.0001}$ & $\underline{0.692}_{\pm0.0001}$ & $0.014_{\pm0.0001}$ & $0.750_{\pm0.0001}$ & $0.007_{\pm0.0001}$ & $0.340_{\pm0.0001}$ & $\underline{0.508}_{\pm0.0001}$ & $0.684_{\pm0.0001}$ & $0.000_{\pm0.0001}$ & -- & $0.000_{\pm0.0001}$ & $0.337_{\pm0.0001}$ \\
\quad TENT \venue{ICLR 2021} & $0.516_{\pm0.002}$ & $0.685_{\pm0.001}$ & $0.022_{\pm0.003}$ & $\underline{0.817}_{\pm0.024}$ & $\underline{0.011}_{\pm0.002}$ & $\underline{0.342}_{\pm0.001}$ & $0.503_{\pm0.001}$ & $\underline{0.688}_{\pm0.003}$ & $\underline{0.111}_{\pm0.003}$ & $\underline{0.779}_{\pm0.030}$ & $\underline{0.059}_{\pm0.002}$ & $\underline{0.365}_{\pm0.001}$ \\
\quad RLCF \venue{ICLR 2024} & $0.516_{\pm0.0001}$ & $0.684_{\pm0.0001}$ & $\mathbf{0.024}_{\pm0.0001}$ & $0.714_{\pm0.0001}$ & $\mathbf{0.012}_{\pm0.0001}$ & $0.341_{\pm0.0001}$ & $0.504_{\pm0.0001}$ & $0.682_{\pm0.0001}$ & $0.042_{\pm0.0001}$ & $0.692_{\pm0.0001}$ & $0.022_{\pm0.0001}$ & $0.345_{\pm0.0001}$ \\
\quad ML-TTA \venue{ICLR 2025} & $0.516_{\pm0.002}$ & $0.685_{\pm0.001}$ & $0.022_{\pm0.003}$ & $\underline{0.817}_{\pm0.024}$ & $\underline{0.011}_{\pm0.002}$ & $\underline{0.342}_{\pm0.001}$ & $0.503_{\pm0.002}$ & $\underline{0.688}_{\pm0.003}$ & $\underline{0.111}_{\pm0.003}$ & $\underline{0.779}_{\pm0.030}$ & $\underline{0.059}_{\pm0.002}$ & $\underline{0.365}_{\pm0.001}$ \\
\rowcolor{oursgray} \quad \textbf{CARVE} & $\mathbf{0.542}_{\pm0.002}$ & $\mathbf{0.719}_{\pm0.001}$ & $\underline{0.023}_{\pm0.003}$ & $\mathbf{0.838}_{\pm0.053}$ & $\underline{0.011}_{\pm0.002}$ & $\mathbf{0.359}_{\pm0.001}$ & $\mathbf{0.528}_{\pm0.002}$ & $\mathbf{0.723}_{\pm0.003}$ & $\mathbf{0.116}_{\pm0.003}$ & $\mathbf{0.798}_{\pm0.003}$ & $\mathbf{0.062}_{\pm0.002}$ & $\mathbf{0.383}_{\pm0.001}$ \\
\addlinespace[2pt]
\multicolumn{13}{l}{\itshape Vocab-FT}\\[1pt]
\quad No-TTA & $0.487_{\pm0.0001}$ & $0.663_{\pm0.0001}$ & $0.631_{\pm0.0001}$ & $0.655_{\pm0.0001}$ & $0.609_{\pm0.0001}$ & $0.528_{\pm0.0001}$ & $0.486_{\pm0.0001}$ & $0.661_{\pm0.0001}$ & $\mathbf{0.195}_{\pm0.0001}$ & $0.681_{\pm0.0001}$ & $\mathbf{0.114}_{\pm0.0001}$ & $\underline{0.377}_{\pm0.0001}$ \\
\quad TENT \venue{ICLR 2021} & $0.495_{\pm0.001}$ & $0.665_{\pm0.001}$ & $0.642_{\pm0.001}$ & $0.659_{\pm0.002}$ & $0.626_{\pm0.003}$ & $0.537_{\pm0.001}$ & $0.496_{\pm0.003}$ & $0.672_{\pm0.002}$ & $0.128_{\pm0.017}$ & $\underline{0.733}_{\pm0.016}$ & $0.070_{\pm0.010}$ & $0.366_{\pm0.006}$ \\
\quad RLCF \venue{ICLR 2024} & $\underline{0.496}_{\pm0.0001}$ & $\underline{0.667}_{\pm0.0001}$ & $\underline{0.649}_{\pm0.0001}$ & $\underline{0.668}_{\pm0.0001}$ & $\underline{0.631}_{\pm0.0001}$ & $\underline{0.547}_{\pm0.0001}$ & $\underline{0.500}_{\pm0.0001}$ & $\underline{0.675}_{\pm0.0001}$ & $0.103_{\pm0.0001}$ & $0.676_{\pm0.0001}$ & $0.056_{\pm0.0001}$ & $0.356_{\pm0.0001}$ \\
\quad ML-TTA \venue{ICLR 2025} & $0.495_{\pm0.001}$ & $0.665_{\pm0.001}$ & $0.642_{\pm0.001}$ & $0.659_{\pm0.002}$ & $0.626_{\pm0.003}$ & $0.537_{\pm0.001}$ & $0.496_{\pm0.003}$ & $0.673_{\pm0.002}$ & $0.126_{\pm0.014}$ & $0.730_{\pm0.011}$ & $0.069_{\pm0.009}$ & $0.366_{\pm0.005}$ \\
\rowcolor{oursgray} \quad \textbf{CARVE} & $\mathbf{0.520}_{\pm0.0001}$ & $\mathbf{0.699}_{\pm0.001}$ & $\mathbf{0.674}_{\pm0.001}$ & $\mathbf{0.692}_{\pm0.002}$ & $\mathbf{0.657}_{\pm0.004}$ & $\mathbf{0.564}_{\pm0.001}$ & $\mathbf{0.521}_{\pm0.003}$ & $\mathbf{0.706}_{\pm0.002}$ & $\underline{0.133}_{\pm0.015}$ & $\mathbf{0.767}_{\pm0.012}$ & $\underline{0.073}_{\pm0.009}$ & $\mathbf{0.384}_{\pm0.005}$ \\
\addlinespace[2pt]
\multicolumn{13}{l}{\itshape Class-FT}\\[1pt]
\quad No-TTA & $0.487_{\pm0.035}$ & $0.675_{\pm0.019}$ & $\underline{0.500}_{\pm0.138}$ & $0.639_{\pm0.018}$ & $\mathbf{0.431}_{\pm0.197}$ & $0.458_{\pm0.044}$ & $0.505_{\pm0.031}$ & $0.678_{\pm0.026}$ & $0.514_{\pm0.021}$ & $0.673_{\pm0.014}$ & $0.416_{\pm0.033}$ & $0.478_{\pm0.002}$ \\
\quad TENT \venue{ICLR 2021} & $\underline{0.545}_{\pm0.003}$ & $0.693_{\pm0.012}$ & $0.486_{\pm0.087}$ & $0.694_{\pm0.003}$ & $0.379_{\pm0.103}$ & $\underline{0.477}_{\pm0.040}$ & $\underline{0.547}_{\pm0.024}$ & $0.699_{\pm0.017}$ & $\underline{0.539}_{\pm0.152}$ & $\underline{0.706}_{\pm0.003}$ & $\underline{0.453}_{\pm0.201}$ & $\underline{0.512}_{\pm0.080}$ \\
\quad RLCF \venue{ICLR 2024} & $0.542_{\pm0.029}$ & $\underline{0.710}_{\pm0.021}$ & $0.375_{\pm0.053}$ & $\underline{0.711}_{\pm0.007}$ & $0.256_{\pm0.050}$ & $0.437_{\pm0.017}$ & $0.531_{\pm0.031}$ & $0.698_{\pm0.019}$ & $0.479_{\pm0.050}$ & $0.703_{\pm0.037}$ & $0.364_{\pm0.048}$ & $0.477_{\pm0.036}$ \\
\quad ML-TTA \venue{ICLR 2025} & $\underline{0.545}_{\pm0.003}$ & $0.693_{\pm0.012}$ & $0.486_{\pm0.087}$ & $0.694_{\pm0.003}$ & $0.379_{\pm0.103}$ & $\underline{0.477}_{\pm0.040}$ & $\underline{0.547}_{\pm0.023}$ & $\underline{0.700}_{\pm0.016}$ & $\underline{0.539}_{\pm0.152}$ & $\underline{0.706}_{\pm0.003}$ & $\underline{0.453}_{\pm0.201}$ & $\underline{0.512}_{\pm0.080}$ \\
\rowcolor{oursgray} \quad \textbf{CARVE} & $\mathbf{0.572}_{\pm0.003}$ & $\mathbf{0.728}_{\pm0.012}$ & $\mathbf{0.512}_{\pm0.090}$ & $\mathbf{0.730}_{\pm0.002}$ & $\underline{0.399}_{\pm0.106}$ & $\mathbf{0.502}_{\pm0.041}$ & $\mathbf{0.574}_{\pm0.025}$ & $\mathbf{0.734}_{\pm0.017}$ & $\mathbf{0.564}_{\pm0.162}$ & $\mathbf{0.740}_{\pm0.005}$ & $\mathbf{0.474}_{\pm0.213}$ & $\mathbf{0.537}_{\pm0.085}$ \\
\bottomrule
\end{tabular}%
}
\end{table*}

\section{Experiments}
\label{sec:experiments}

\subsection{Setup}
\label{sec:setup}

\textbf{Models and datasets.}
We evaluate two 3D CT VLM backbones trained on CT-RATE. CT-CLIP~\cite{hamamci2024ctclip} produces a global volume representation and is tested in three variants that share the architecture but differ in the fine-tuning stage: Zero-shot, Vocab-FT, and Class-FT. fVLM~\cite{shui2025fvlm} is an anatomy aware model with fine-grained image-text alignment. Our primary setting is multi-label abnormality classification, evaluated internally on the CT-RATE validation set and externally on RAD-ChestCT. To test whether the conclusions depend on label count, we also evaluate CC-CCII in a three class setting and LUNA16 in a binary setting. Dataset sizes, patient level splits, label harmonization, prompt templates, and F1/accuracy threshold selection are in Appendix~\ref{app:repro}.

\noindent\textbf{Methods and protocol.}
We compare No-TTA, TENT~\cite{wang2021tent}, RLCF~\cite{zhao2024test}, ML-TTA~\cite{wu2025mltta}, and CARVE under the same episodic protocol. Each test scan is adapted independently, with the model reset before the next scan. Adaptation updates only selected visual normalization parameters using Adam with learning rate $10^{-5}$ for two steps. CARVE uses $V{=}8$ weak 3D views, retains the lowest entropy $25\%$, estimates $\khat$ by Eq.~\ref{eq:khat}, and optimizes Eq.~\ref{eq:carve} with $\lambda_{\mathrm{neg}}{=}0.8$. Hyperparameters are fixed once and reused unchanged for CT-RATE validation
and all external cohorts. No target labels are used for model or hyperparameter selection. Volumes are HU clipped to $[-1000,1000]$, resampled to $(1.5,0.75,0.75)$\,mm, and evaluated at native depth $z{=}240$ unless stated.

\noindent\textbf{Reporting.}
Tables~\ref{tab:main-ctclip-ctrate-internal}-\ref{tab:gen-ctclip-luna16-2-labels}
report $\text{mean}_{\pm\sigma}$ over independent test-time adaptation runs with different adaptation seeds. The standard deviation reflects adaptation stability under stochastic inference-time perturbations. Conclusions are based on trends that remain consistent across architectures, model variants, cohorts, and evaluation metrics.

\subsection{Main results}
\label{sec:main}

\textbf{Volumetric integrity determines TTA effectiveness.}
\Cref{tab:main-ctclip-ctrate-internal} shows that input depth is not a benign implementation choice. On internal CT-RATE, the unadapted CT-CLIP Zero-shot version reaches AUROC $0.713$ at native depth $z{=}240$, but drops to $0.588$ at $z{=}40$, a decrease of $0.125$ with input depth as the only change. The reduced depth model also shifts toward overprediction, with recall $0.914$ and precision $0.198$. The same pattern appears across other variants, with Vocab-FT dropping from $0.763$ to $0.654$ and Class-FT dropping from $0.520$ to $0.506$. On RAD-ChestCT, however, the unadapted AUROC is near chance at both depths, $0.499$ at $z{=}240$ and $0.511$ at $z{=}40$ (\Cref{tab:main-ctclip-rad-external}). Thus, native depth repairs the internal zero-shot base, but it does not remove external representational shift. We therefore report all main results at native depth and treat the external benchmark as the real test of adaptation under shift.

% \noindent\textbf{Entropy-based adaptation mainly moves the operating point.}

\noindent{\textbf{Entropy TTA exposes the multi-label objective mismatch.}}
\Cref{tab:main-ctclip-ctrate-internal,tab:main-ctclip-rad-external,tab:main-fvlm-sidebyside}
summarize the abnormality results. The entropy-based objectives behave consistently with the mismatch discussed in Sec.~\ref{sec:multilabel}. On
internal CT-RATE Zero-shot, TENT increases recall from $0.698$ to $0.712$ but reduce accuracy from $0.633$
to $0.605$, indicating that they sharpen confident labels without estimating the scan-level positive count. RLCF moves the operating point
in the opposite direction, reducing recall to $0.682$. Under the stronger external shift of RAD-ChestCT, these objectives produce only small AUROC
changes, consistent with a regime in which the base predictions contain little separable signal. 
% These patterns motivate estimating a scan-level cardinality rather than treating prediction as either a single categorical distribution or a set of unrelated binary decisions.

\noindent\textbf{CARVE gains most when the base already contains usable structure.}
On CT-CLIP, CARVE gives the strongest AUROC trend across variants and also improves the balance between recall and precision. On internal CT-RATE Zero-shot, it reaches AUROC $0.749$ compared with $0.713$ for No-TTA. On external RAD-ChestCT Zero-shot, it improves AUROC from $0.499$ to $0.533$. These gains support the role of scan specific cardinality and retained view selection, but they should be read as operating point improvements when the base model already carries usable structure, not as evidence that adaptation can create discrimination from a failed representation.

\noindent\textbf{The base model gap is larger than the adapter gap.}
The strongest effect under external shift is the base representation. On RAD-ChestCT, unadapted fVLM reaches AUROC $0.612$, while unadapted CT-CLIP is near chance at $0.499$, a $+0.11$ gap that no CT-CLIP adaptation objective closes. CARVE also improves fVLM, reaching AUROC $0.642$, but the base model gap remains the dominant deployment lever. 
% This supports the central conclusion: TTA is conditional rather than automatic. A cardinality aware objective helps in the favorable regime, while severe external representational shift requires a stronger transferred base model.
This confirms the boundary identified in \Cref{sec:diagnosis}: under severe external shift, adaptation is limited by the transferred base representation.

\noindent\textbf{Transfer across smaller label spaces.}
Tables~\ref{tab:gen-ctclip-ccccii-3-labels} and~\ref{tab:gen-ctclip-luna16-2-labels}
apply the same objectives to CC-CCII and LUNA16. CARVE applies without modification as the label set becomes smaller; in the binary case, the cardinality estimate reduces to a single label presence decision. The pattern remains consistent: when the zero-shot base is near chance, adaptation mainly shifts the operating point, while stronger movement appears only when the model already separates the classes. We therefore read these results as evidence that CARVE transfers across classification settings, not as evidence that any objective can recover discrimination when the base lacks it.

\begin{figure}[t]
  \centering
  \includegraphics[width=\linewidth]{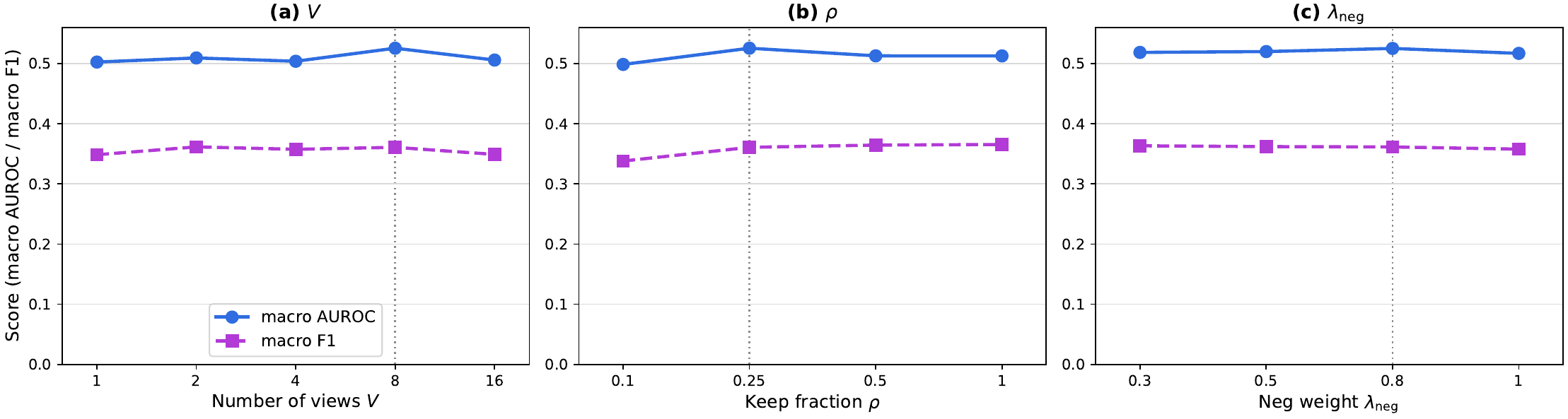}
  \caption{\textbf{Hyperparameter sensitivity.}
  CT-CLIP on RAD-ChestCT. CARVE is stable around the default setting
  $V{=}8$, $\rho{=}0.25$, and $\lambda_{\mathrm{neg}}{=}0.8$.}
  \label{fig:hparam}
\end{figure}

\begin{figure}[t]
  \centering
  \includegraphics[width=1\linewidth]{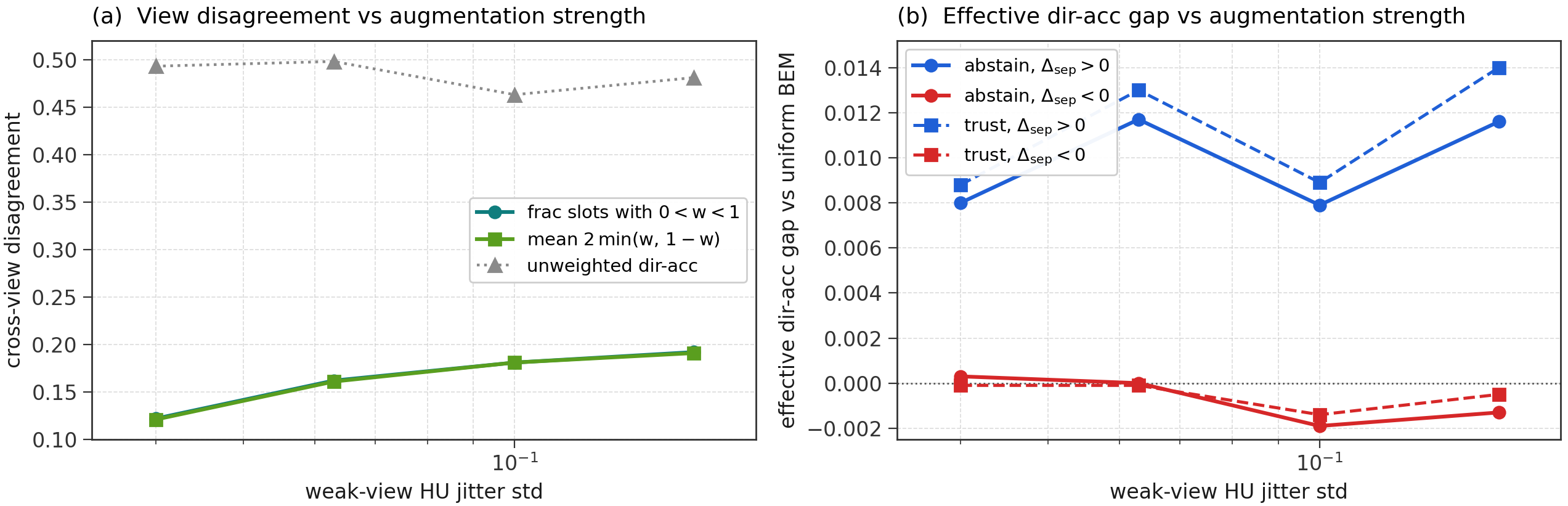}
  \caption{\textbf{Weak view strength.} Stronger perturbations raise disagreement but do not improve adaptation, motivating milder distortions.}
  \label{fig:augsweep}
\end{figure}

\begin{figure}[t]
  \centering
  \includegraphics[width=\linewidth]{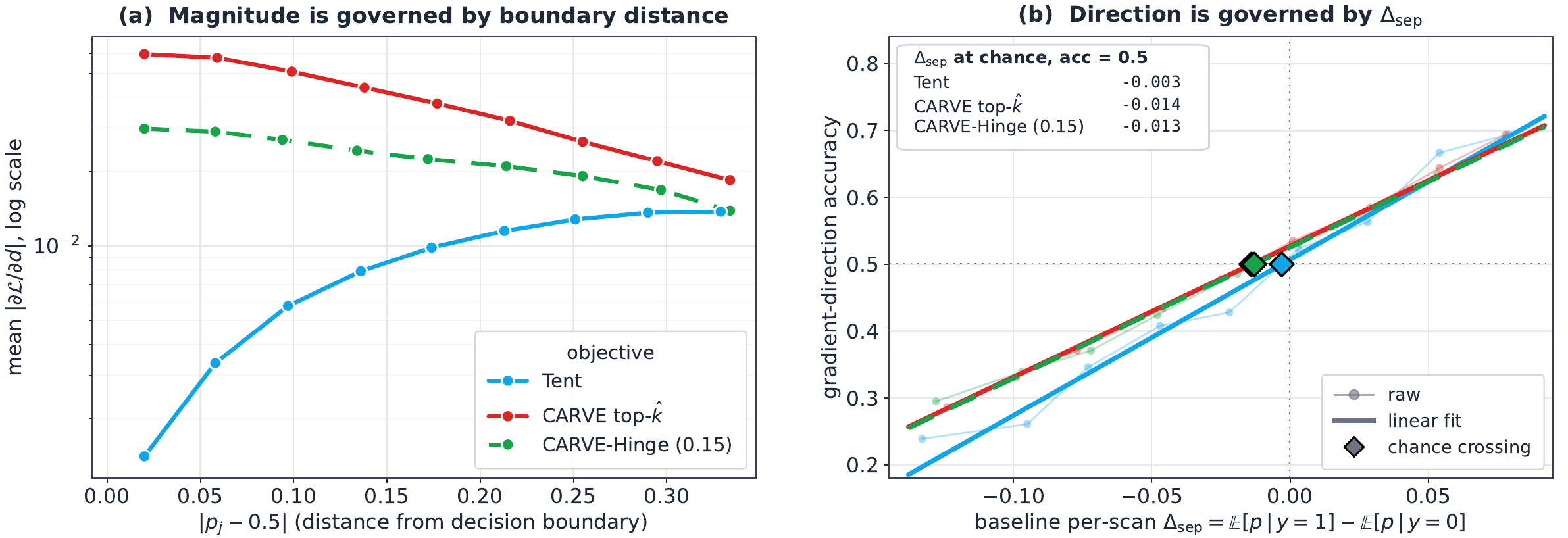}
%   \caption{\textbf{TTA failure axes.}
% Entropy loses gradient magnitude near the boundary, while top-$\hat{k}$
% objectives retain it. However, direction accuracy depends on baseline
% separation and collapses when the base model has no signal.}
\caption{\textbf{TTA failure axes.} Top-$\hat{k}$ objectives keep gradient magnitude near the boundary where entropy fades, but direction still fails when the base lacks usable structure.}
  \label{fig:graddiag}
\end{figure}

\begin{figure}[t]
  \centering
  \includegraphics[width=\linewidth]{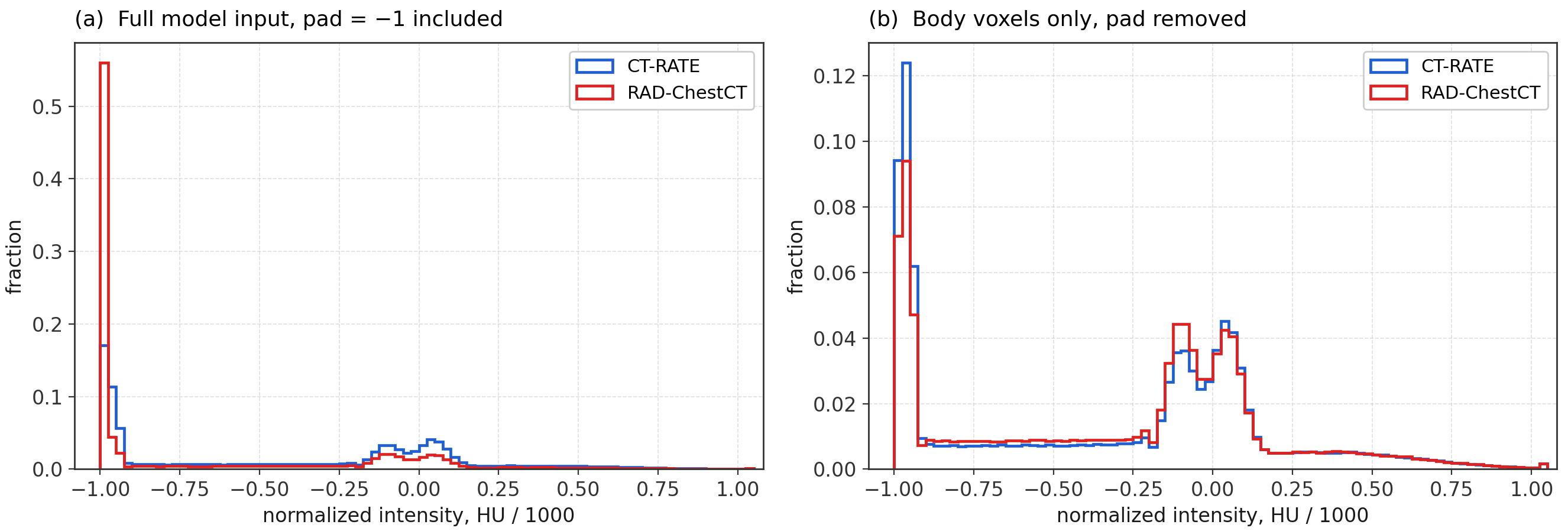}
  \caption{\textbf{Input intensity comparison.} Internal and external inputs share similar intensity distributions, so the external deficit is representational rather than a preprocessing artifact.}
  \label{fig:inputhist}
\end{figure}

\begin{figure*}[t]
  \centering
  \includegraphics[width=0.32\linewidth]{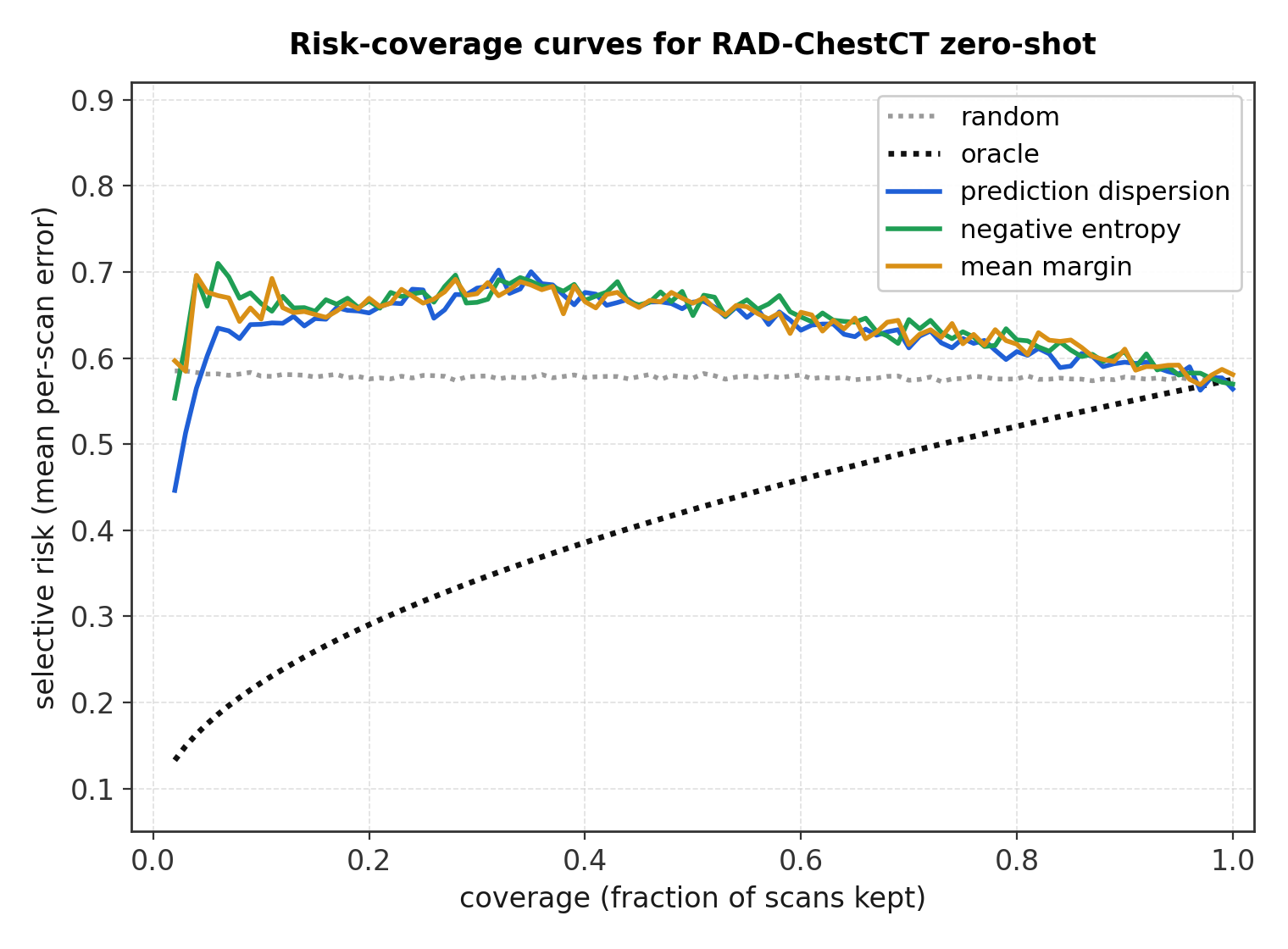}\hfill
  \includegraphics[width=0.32\linewidth]{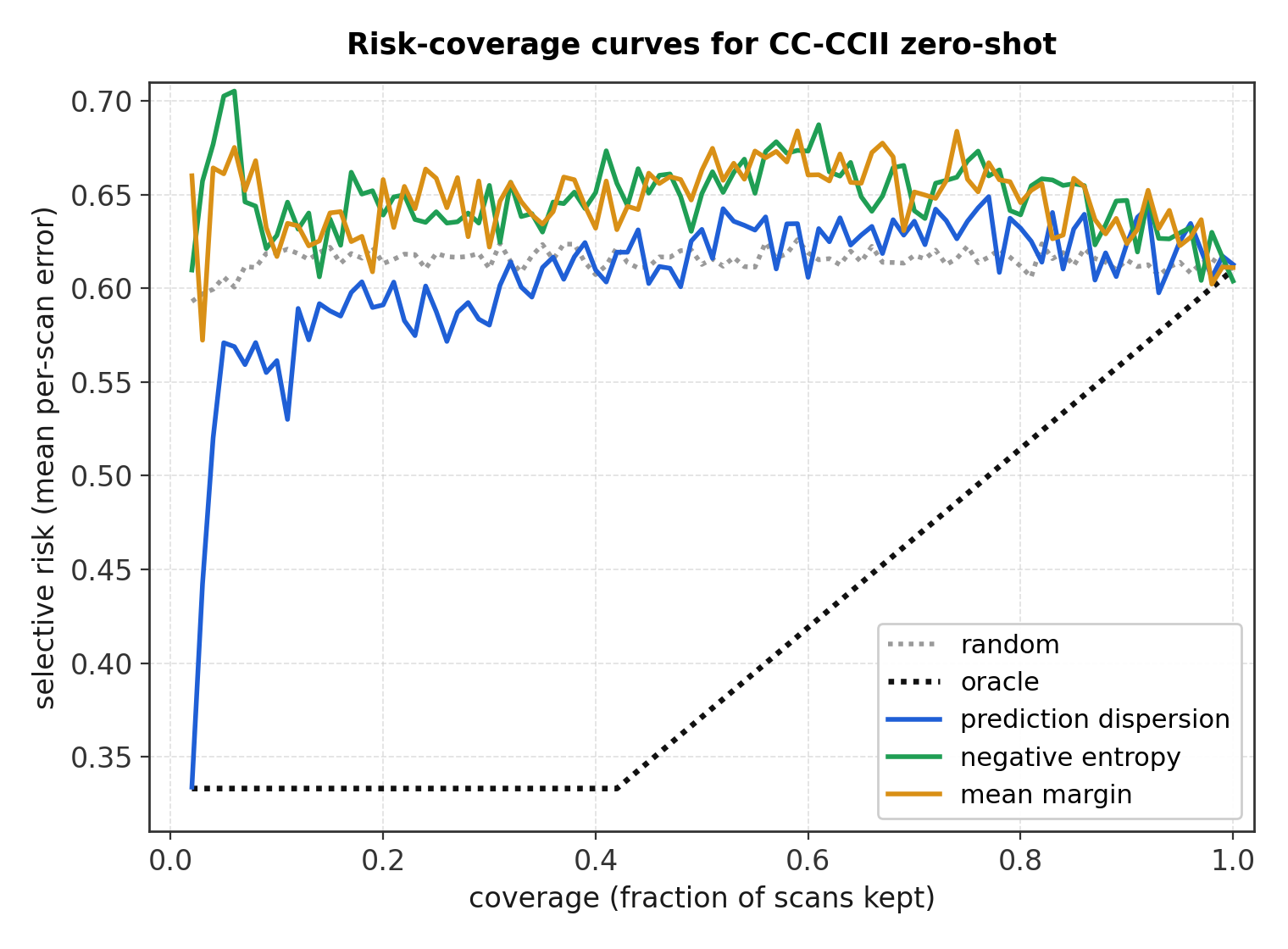}\hfill
  \includegraphics[width=0.32\linewidth]{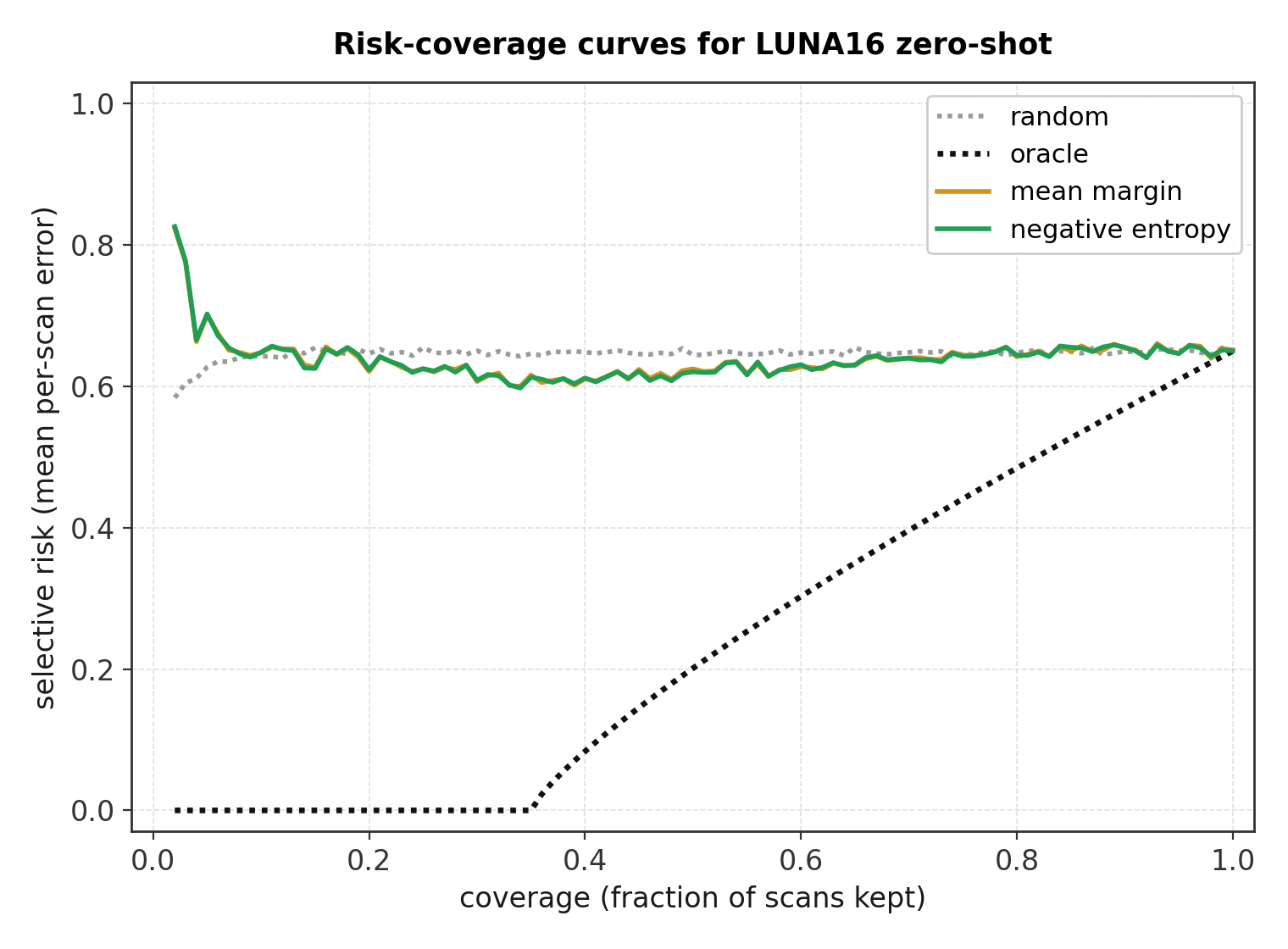}
  \caption{\textbf{Risk coverage of the dispersion gate.} $\sigma_{\mathrm{pred}}$ falls far short of the oracle; the gate only abstains under collapsed predictions to avoid wasteful updates, not detecting per-scan viability.}
  \label{fig:riskcov}
\end{figure*}

\subsection{Ablations and Analysis}
\label{sec:ablation}

\noindent\textbf{Component sensitivity.}
\label{sec:ablation-components}
Figure~\ref{fig:hparam} sweeps the three CARVE hyperparameters on RAD-ChestCT. Performance is stable across a broad range and peaks near defaults used throughout: $V{=}8$ views, keep fraction $\rho{=}0.25$, and negative weight $\lambda_{\mathrm{neg}}{=}0.8$. More views help until saturation, an intermediate keep fraction outperforms keeping all views or only one, and a moderate negative weight balances sharpening the estimated positive labels with suppressing the complement.

Figure~\ref{fig:augsweep} tests the weak view generator. Increasing augmentation strength increases cross view disagreement, but does not improve agreement with the useful update direction. CARVE therefore uses mild HU jitter and small spatial shifts: the multi view step is for reliability filtering, not for strong data augmentation.

\noindent\textbf{Mechanism and gating.}
\label{sec:ablation-mechanism}
Figure~\ref{fig:graddiag} separates two failure axes. First, entropy gradients weaken near $p{=}0.5$, while the top-$\khat$ objective retains update magnitude near the decision boundary. This is the part CARVE addresses. Second, the direction of the update depends on whether the base predictions already separate the classes. When baseline separation is near zero, all objectives have near chance direction accuracy. Thus CARVE can realize available signal, but no objective can supply a correct direction when the base model does not contain one.

Figure~\ref{fig:inputhist} checks whether external deficit is caused by a simple input mismatch. The model input intensity distributions for CT-RATE and RAD-ChestCT are closely aligned, both with and without background padding. Since the external scores still collapse, the deficit is better explained as a representational shift than as a preprocessing artifact.

Finally, Fig.~\ref{fig:riskcov} evaluates the dispersion gate. The per scan dispersion $\sigma_{\mathrm{pred}}$ is a weak selective prediction signal: it is near random on RAD-ChestCT and LUNA16, modest on CC-CCII, and far from oracle performance. We therefore use it only as a conservative abstention signal under collapsed predictions, not as a detector of scans where adaptation is guaranteed to help.

\section{Conclusion}
\label{sec:conclusion}

We studied \emph{when} TTA helps zero-shot 3D CT VLMs rather than assume it always
does. Adaptation is conditional. It requires a depth-consistent input pipeline and
a base representation that transfers to the target cohort, so that prompt-pair
predictions separate present from absent abnormalities. CARVE is built for this
regime, estimating a scan-specific positive-label cardinality and preserving
co-occurring abnormalities through memory-efficient multi-view updates on retained
weak 3D views. It gives the most consistent improvements across CT-CLIP, fVLM, and
the multi-label, three-class, and binary settings when the base model is already
discriminative, but under severe external shift base-model transfer remains the
dominant factor. We therefore frame multi-label TTA for zero-shot 3D CT VLMs as a
distinct problem whose value depends on depth-consistent input, base transfer, and
a cardinality-aware objective for prompt-pair multi-label prediction.

{
    \small
    \bibliographystyle{ieeenat_fullname}
    \bibliography{main}
}

\clearpage
\onecolumn
\appendix

\section{Implementation details}
\label{app:implementation}

\paragraph{Adapted parameters.}
CARVE adapts only the affine scale and shift of the visual normalization
layers in the selected transformer blocks of the image encoder, following the
norm-only protocol of Tent~\cite{wang2021tent}. All other visual parameters and
all text-side parameters remain frozen. Adaptation is episodic: for every test scan, the normalization parameters are reset to their pretrained values, updated
for two gradient steps with Adam at learning rate $10^{-5}$, and then used to
score the unperturbed volume. The set of adapted blocks is fixed on the internal
CT-RATE validation split and reused for every backbone, cohort, and variant, Appendix~\ref{app:repro}.

\paragraph{Cardinality bounds.}
The scan-specific cardinality in \Cref{eq:khat} is clipped to
$[k_{\min},k_{\max}]$. We set $k_{\min}{=}1$, so that at least one abnormality is
preserved and the positive set is never empty, and
$k_{\max}{=}8$, which caps the positive set below the label-space
size $L$ to prevent noisy or poorly calibrated probabilities from selecting an
implausibly large positive set. Both bounds are fixed across all cohorts and are
not tuned per dataset.

\paragraph{Weak-view generation.}
The $V{=}8$ weak views are produced by composing two mild volumetric
perturbations at the reference strength: additive Gaussian intensity jitter
($\sigma{=}0.02$ on the volume normalized by $1000$, $\approx 20$\,HU) and an
integer spatial translation of up to $\pm3\%$ of each axis (circular shift). No
crops, resizes, flips, or rotations are used. The first view ($v{=}0$) is the
unperturbed volume. Augmentation strength is deliberately mild; \Cref{fig:augsweep} shows
that stronger perturbations raise cross-view disagreement without improving the
update direction, so the multi-view step serves reliability filtering rather
than data augmentation.

\paragraph{Dispersion gate.}
The optional dispersion gate compares the per-scan prediction dispersion
$\sigma_{\mathrm{pred}}(x)$ defined in \Cref{sec:gatedef} against a fixed
threshold $\tau{=}0.091$, abstaining from adaptation when
$\sigma_{\mathrm{pred}}(x)\le\tau$. The threshold is selected on the internal
validation split and held fixed for all external cohorts; it is used only as a
conservative abstention rule, not as a per-scan viability detector
(\Cref{sec:ablation}).

\paragraph{Memory-efficient update.}
CARVE uses a two-stage update to keep volumetric adaptation within memory. It
first scores all $V$ weak views without gradients to obtain the per-view
entropies, the retained-view average $\bar{\mathbf{p}}$, and the cardinality
estimate $\khat$. It then recomputes only the $K$ retained views with gradients
for the adaptation step, so gradient activations are stored for $K$ rather than
$V$ views. All experiments use batch size $1$ on a single NVIDIA L40S.

\section{Reproducibility and Experimental Detail}
\label{app:repro}

\paragraph{Compute and preprocessing.}
All experiments use batch size 1 on a single NVIDIA L40S. Input CT volumes are
loaded from NPZ, converted to Hounsfield Units when rescale slope/intercept are
available, clipped to $[-1000,1000]$, normalized by $1000$, resampled to
$(1.5,0.75,0.75)$\,mm along $(z,x,y)$, and run at native depth $z{=}240$
($40\times480\times480$ at $z{=}40$). An image-once encoding caches the volume
embedding so that scoring $L$ prompt pairs does not re-encode the image, which
is what keeps native-depth inference feasible.

\paragraph{Datasets and splits.}
The backbones are pretrained on CT-RATE; we evaluate in-distribution on its
held-out validation split, since CT-RATE provides no separate test split.
RAD-ChestCT is the external 16-abnormality benchmark, CC-CCII is a three-class
COVID cohort, and LUNA16 (LIDC-IDRI) is the binary nodule benchmark. All splits
are \emph{patient-level} with no subject shared across splits. CT-RATE comprises
$20{,}000$ training patients ($\approx 47{,}149$ reconstructions) and $1{,}564$
validation patients ($3{,}039$ reconstructions); The external and generalization
cohorts contribute $360$ scans (RAD-ChestCT), $135$ scans (CC-CCII), and $621$
scans (LUNA16).

\paragraph{Label harmonization.}
CT-CLIP scores CT-RATE on its native $18$-label set and RAD-ChestCT on its
native $16$-label set; the external comparison uses the correspondence between
them. Of these, $14$ abnormalities map one-to-one. RAD-ChestCT's single
\emph{Calcification} corresponds to two CT-RATE labels, \emph{Arterial wall
calcification} and \emph{Coronary artery wall calcification}; RAD-ChestCT's
\emph{Hietal Hernia} corresponds to CT-RATE's \emph{Hiatal hernia}; and CT-RATE's
\emph{Mosaic attenuation pattern} has no RAD-ChestCT counterpart. CC-CCII and
LUNA16 use their native label sets unchanged.

\paragraph{Prompt templates.}
Each abnormality $j$ is scored with a positive/negative prompt pair whose phrasing
differs by backbone (see \Cref{fig:overview}). CT-CLIP uses ``There is
\{abnormality\}.'' and ``There is no \{abnormality\}.''; fVLM uses ``there is
absolutely \{abnormality\} present.'' and ``there is absolutely not
\{abnormality\} present.'' Text-side parameters are frozen throughout.

\paragraph{Thresholds for F1/accuracy.}
AUROC and AUPRC are threshold-free. F1, precision, recall, and accuracy use a
fixed threshold of $0.5$ on the per-label presence probability $p_j$, applied
identically to every method so that operating-point comparisons are fair.

\paragraph{Hyperparameter selection.}
We fixed $V=8$, $\rho=0.25$, $\lambda_{\mathrm{neg}}=0.8$,
learning rate $10^{-5}$, two adaptation steps, and the adapted normalization
blocks using internal CT-RATE validation only, and reused these settings
unchanged for every backbone, cohort, and checkpoint. No target labels were
used for selection. Figure~\ref{fig:hparam} reports a post-hoc sensitivity
sweep on RAD-ChestCT around the fixed defaults to assess robustness, not to
choose hyperparameters. The external and generalization cohorts are used only
for test-time evaluation.

\paragraph{fVLM evaluation.}
We use the released fVLM weights in their native mask-free inference mode and
apply the same five objectives to the same normalization-style parameters as for
CT-CLIP. Metric definitions are identical to CT-CLIP (prompt-pair two-class
softmax giving per-label Bernoulli probabilities, macro-averaged). Image features
are the mean over the four organ-query vision projections; volumes use fVLM's
native window $[-1150,350]\to[0,1]$ and trilinear resize to
$112\times256\times352$, and adaptation updates the $50$ visual-encoder LayerNorm
tensors ($38{,}400$ parameters). Because the official organ-ROI segmentation crop
is omitted (masks unavailable), we do not claim parity with the fVLM paper's
reported zero-shot numbers.

\paragraph{Reported variability.}
Tables~\ref{tab:main-ctclip-ctrate-internal}-\ref{tab:gen-ctclip-luna16-2-labels}
report mean$\pm\sigma$ over repeated test-time runs. For stochastic adaptation
methods, runs use different adaptation seeds. The standard deviation reflects
run-to-run stability under stochastic inference-time distortions, not statistical
uncertainty over patients.

\end{document}